\documentclass[conference]{IEEEtran}
\usepackage{times}

\usepackage{amsmath}
\usepackage{amssymb}
\usepackage{amsthm}
\usepackage[numbers]{natbib}
\usepackage{multicol}
\usepackage[bookmarks=true]{hyperref}
\usepackage{hyperref}
\usepackage{cleveref}
\usepackage{booktabs}
\usepackage{graphicx} 
\usepackage{tabularx}
\usepackage{array}
\newcolumntype{Y}{>{\centering\arraybackslash}X}
\usepackage{array}
\usepackage{multirow}
\let\labelindent\relax 
\usepackage{enumitem}
\usepackage{listings}
\usepackage{xcolor}
\usepackage[frozencache,cachedir=.]{minted}
\usepackage{xspace}

\usepackage[english]{babel}
\usepackage[utf8]{inputenc}
\usepackage{algorithm}
\usepackage[noend]{algpseudocode}
\usepackage{mathtools}

\usepackage{bm}
\usepackage{wrapfig}
\usepackage{tikz}
\usepackage{pgfplots}
\pgfplotsset{compat=newest}
\usepackage{verbatim}
\usepackage{flushend}
\usepackage[font={footnotesize}]{caption}
\usepackage{subcaption}

\newcommand{\bs}{\bm{s}}

\newcommand{\bF}{\bm{F}}

\newcommand{\bJ}{\bm{J}}

\newcommand{\cJ}{\mathcal{J}}

\newcommand{\cS}{{\mathcal{S}}}

\newcommand{\cU}{\mathcal{U}}

\newcommand{\cX}{\mathcal{X}}

\newcommand{\II}{\mathbb{I}}

\newcommand{\RR}{\mathbb{R}}

\newcommand{\bphi}{\bm{\phi}}

\newcommand{\bETA}{\bm{\eta}}

\newcommand{\minimize}{\mathop{\mathbf{min}}}

\newcommand{\minimizewrt}[1]{\mathop{\underset{#1}{\minimize}}}

\newcommand{\defeq}{\vcentcolon=}

\newcommand{\st}{\mathop{\mathbf{s.t.}}}

\newtheorem{problem}{Problem}

\newtheorem*{remark*}{Remark}

\colorlet{revision_color}{black}

\begin{document}

\title{Dexterous Safe Control for Humanoids in Cluttered Environments via Projected Safe Set Algorithm}


\author{
\authorblockN{Rui Chen, Yifan Sun, and Changliu Liu}
\authorblockA{\textit{Robotics Institute, Carnegie Mellon University}\\
\textit{\{ruic3, yifansu2, cliu6\}@andrew.cmu.edu}}
}


%

\twocolumn[{%
\renewcommand\twocolumn[1][]{#1}%
\maketitle
\begin{center}
    \centering
    \captionsetup{type=figure}
    \includegraphics[width=\textwidth]{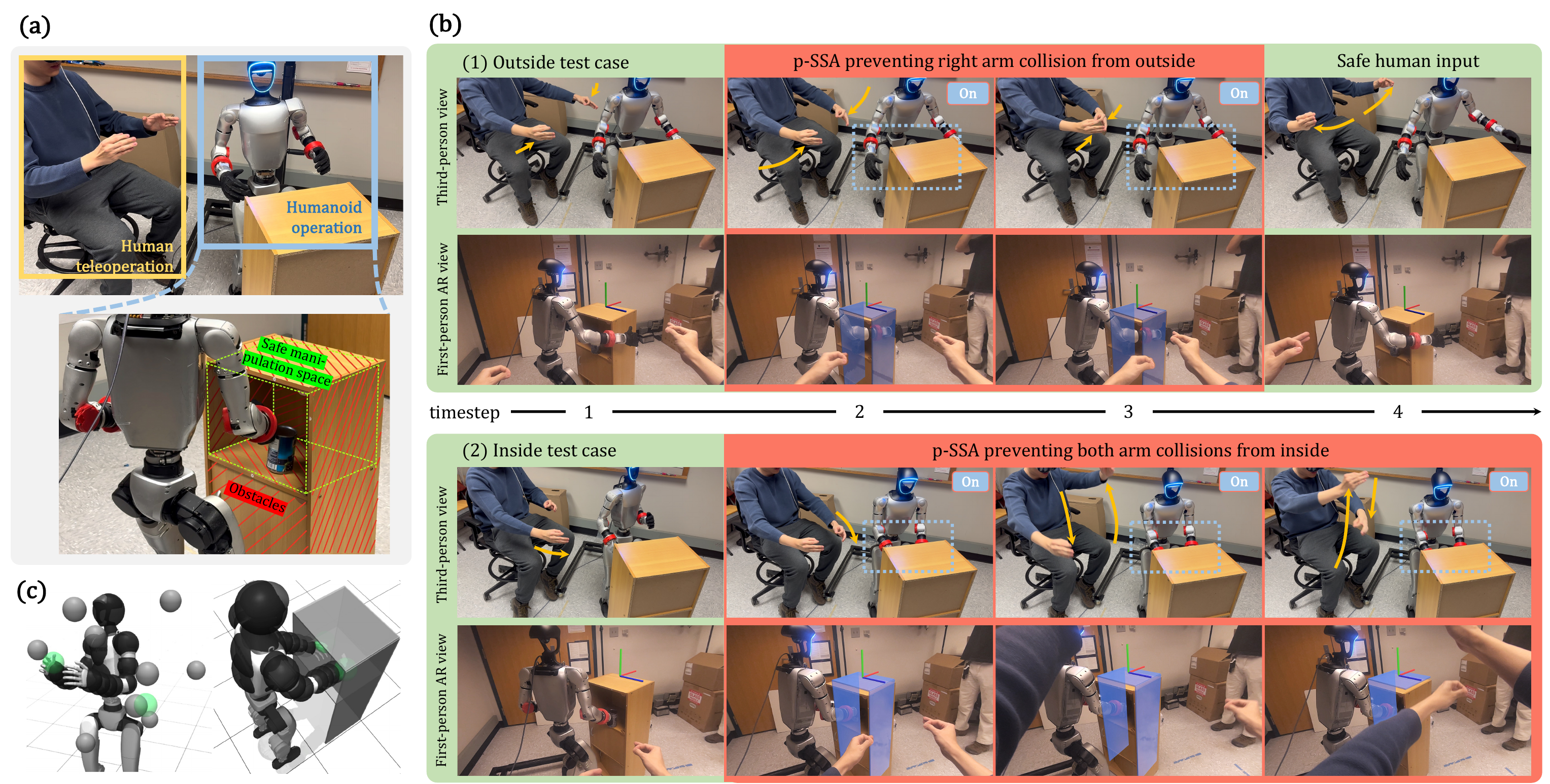}
    \captionof{figure}{
    Application of dexterous safe control for humanoids in cluttered environments.
    (a) A safe teleoperation task where the Unitree G1 humanoid mirrors Apple Vision Pro (AVP)-tracked human arm motions to manipulate objects inside a cabinet while avoiding both obstacle and self-collisions.
    (b)-(1) p-SSA blocks excessive squeezing near the cabinet (frames 2–3) and allows safe input in frame 4.
    Obstacles triggering p-SSA intervention are highlighted in blue in AVP view.
    (b)-(2) p-SSA prevents collisions when both arms are inside the cabinet.
    This poses a more complex problem due to multiple concurrent safety constraints within a confined space.
    (c) Simulated goal-reaching (green) with collision avoidance with multiple obstacles (gray) via p-SSA.
    }
    \label{fig:teaser}
\end{center}%
}]

\begin{abstract}

It is critical to ensure safety for humanoid robots in real-world applications without compromising performance.
In this paper, we consider the problem of dexterous safety, featuring limb-level geometry constraints for avoiding both external and self-collisions in cluttered environments.
Compared to safety with simplified bounding geometries in sprase environments, dexterous safety produces numerous constraints which often lead to infeasible constraint sets when solving for safe robot control.
To address this issue, we propose Projected Safe Set Algorithm (p-SSA), an extension of classical safe control algorithms to multi-constraint cases.
p-SSA relaxes conflicting constraints in a principled manner, minimizing safety violations to guarantee feasible robot control.
We verify our approach in simulation and on a real Unitree G1 humanoid robot performing complex collision avoidance tasks.
Results show that p-SSA enables the humanoid to operate robustly in challenging situations with minimal safety violations and directly generalizes to various tasks with zero parameter tuning.

\end{abstract}

\IEEEpeerreviewmaketitle

\section{Introduction}\label{sec:intro}


Humanoid robots are increasingly being deployed in the real world.
Their highly articulated physical structures grant them remarkable dexterity to operate even in cluttered environments; those environments are commonly seen in tasks from factory automation to service and healthcare applications.
Yet, as their capabilities grow, ensuring the safety of the humanoid and its environment becomes both critical and difficult.
A common approach to collision avoidance is to simplify the robot’s geometry—for instance, by wrapping the entire body using a single bounding cylinder \cite{yun2024safe, he2024agile}.
While such geometric reductions enable straightforward algorithms and help in early proof-of-concept implementations, they tend to be overly conservative for humanoid robots, severely restricting feasible motion and task performance.
Realizing the full potential of humanoids calls for more precise and fine-grained modeling of the geometry of both the robot and the environment, especially for tasks involving close-proximity interactions.
We refer to the problem of enforcing safety under limb-level geometry modeling as \textit{dexterous safety}.
In this paper, we aim to address dexterous safety in cluttered environments without compromising performance.




Regarding approaches to safe control, researchers have broadly pursued two classes of methods: indirect (model-based) and direct (model-free).
Indirect approaches such as safe set algorithms (SSA)  \cite{liu2014control, chen2023sis}, control barrier functions (CBFs) \cite{ames2016control, xiao2019control}, and Hamilton–Jacobi (HJ) reachability \cite{choi2021robust} explicitly model the robot dynamics and derive safe control laws accordingly.
They allow interpretable decompositions of objectives and constraints and can be adapted to modified problems (e.g., changed safety criteria) without a fine-tuning process.
In contrast, direct approaches such safe reinforcement learning bypass robot modeling and incorporate tasks with constraints into a single optimization framework.
Despite their promise, direct methods often require extensive training for high-dimensional problems and retraining when new constraints are introduced.
This paper focuses on indirect methods to accommodate the high dimensionality and flexibility of dexterous safety in cluttered environments.


Indirect methods are normally energy function-based approaches \cite{wei2019unified} that (a) synthesize an energy function that quantifies safety and (b) derive a control constraint from that energy function to enforce safety.
The difficulty of applying indirect methods can be characterized by two key aspects: the granularity of robot geometry modeling and the distribution of obstacles.
The robot geometry can be modeled either coarsely using a single convex bounding shape such as a box or cylinder (i.e., \textbf{single-body}), or at limb level with each movable link enclosed with an independent shape (i.e., \textbf{multi-body}).
The surrounding obstacles can be distributed either \textbf{sparsely} such that the robot deals with one obstacle at a time, or \textbf{densely} such that the robot has to jointly consider multiple potential collisions.
Prior works mostly focus on the simplest setting with a single rigid body in sparse environments \cite{yun2024safe, he2024agile, singletary2022onboard, pandya2024multimodal, molnar2021model, zhao2021zeroviolation, choi2023constraint}.
Some work could handle single-body safety in dense environments~\cite{chen2021safe, dawson2022safe, zheng2022clustered} by invoking one safety constraint for each surrounding obstacle. However, these approaches are limited to simple 2D problems and do not come with any safety guarantees. 
Regarding multi-body safety, existing literature mostly assumes sparse environments and solves these problems by reducing them to maximizing the closest distance from the robot to the obstacle which could be regulated using one single energy function \cite{liu2023proactive, lin2017real, liu2022safe}.

Our problem of dexterous safety for humanoids in cluttered environments falls into the most difficult category: multi-body safety with densely distributed obstacles.
In all three cases above with either single body or sparse obstacles, the safety is quantified with respect to one point on the robot body. In the case of single-body safety, it is quantified with respect to the center of the robot; in the case of multi-body safety, it is quantified with respect to the closest point to the obstacle. Hence only one energy function is needed for the robot. For relatively sparse environments with high safe control frequency, this strategy works for multi-body safety as the clearance is high and safety hazards could be mitigated in time. 
However, the complex interaction between multiple robot links and multiple obstacles in proximity can hardly be treated in the same way.
That is because some motion that decreases this one energy function (e.g., by moving the closest point away from the obstacle) may result in another part on the robot body immediately colliding with another obstacle. 
To the best of the authors' knowledge, there is no existing work that can reliably synthesize one single energy function to achieve high safety performance when multiple collisions with multi-body robots are possible. 
We hypothesize that multi-body safety in dense environments is indeed a \textit{multi-objective optimization problem} which can hardly be captured by a single energy function as one single performance index. 
Hence, we must deploy \textit{multiple energy functions} to more precisely capture the safety.
With each energy function leading to a control constraint, we resort to multi-constraint safe control approaches in this paper.

The aforementioned safety approaches (SSA, CBF, etc.) come with nice theoretical safety guarantees such as forward invariance \cite{liu2014control, chen2023sis, ames2016control}.
They often derive a linear safe control constraint with a quadratic objective (e.g., nominal control tracking), forming quadratic programming (QP) problems to solve online.
In our case, this naturally extends to a QP with multiple constraints.
The solution (if it exists) to this QP would satisfy each control constraint and inherit all safety guarantees of the single-constraint version.
However, as will be shown in this paper, multi-constraint QPs can frequently become infeasible either due to inherently/physically infeasible problems or incompatible energy functions (e.g., derived control constraints), unless certain safety requirements are relaxed (e.g., temporarily allowing contact).
This is because, in dexterous safety, we need to carefully constrain the motion of each rigid body in the robot kinematics chain, which is very different from safety for a single rigid body.
That leads to a highly restricted solution space with or without bounded control.
Several previous works consider multi-constraint cases but without a focus on QP feasibility, either due to inherently consistent constraints describing non-trivial connected and closed safe zones \cite{djeha2023robust, nguyen20163d} or large actuation limits \cite{khazoom2022humanoid}.
There are works explicitly avoiding QP infeasibility via Safety Index Synthesis but only focus on a single safety constraint, mostly with problem dimensions no more than five \cite{zhao2023sos, chen2023sis, chen2023sia, liu2022inputsat}.
\cite{breeden2023compositions} composes multiple constraints into a feasible controller but suffers poor scalability; it only shows success on a four-dimensional problem with two constraints, which takes over 1000 seconds to compute.
In the task to be considered in this paper, we model a 29-DoF Unitree G1 humanoid with 19 collision volumes, with more than 10 obstacles in proximity at a time in cluttered environments.
Considering self-collision, our QP contains more than 200 constraints in dexterous safety tasks.
There does not exist any method that can synthesize multiple compatible energy functions to make the QP persistently feasible for high-dimensional problems like ours.

In summary, for dexterous safety for humanoids in cluttered environments, a single energy function can hardly handle multiple safety objectives, while no existing approach could synthesize multiple energy functions that are compatible to guarantee feasible controls.
We argue that under the currently available theoretical tools, the closest to what we desire is to design multiple energy functions for each collision pair of interest and force a multi-constrained QP.
Instead of aiming for guarantees, we desire \textit{a practical method that can be quickly deployed on humanoids with minimal violations of the control constraints when the QP becomes infeasible}, either due to inherently infeasible situations or incompatible energy functions.
To this end, we propose the Relaxed Safe Set Algorithm (r-SSA), which relaxes infeasible control constraints with weighted slack regularization.
We then introduce the Projected Safe Set Algorithm (p-SSA), which improves over r-SSA by removing parameter tuning via decoupled optimization for feasibility and task objectives.
Validation in both simulation and hardware shows that r-SSA can readily compute safe control in challenging dexterous safety tasks with several hundreds of constraints.
At the same time, p-SSA gains top performance across a variety of tasks with zero parameter tuning.
Our contributions can be summarized as follows.
\begin{itemize}
    \item We introduce a novel task of dexterous safety in cluttered environments and analyze the challenges faced by safe control approaches regarding infeasibility from multiple sources.
    \item We propose the Projected Safe Set Algorithm (p-SSA), a novel safe control method for dexterous safety that relaxes conflicting control constraints with minimal violations.
    \item We compare p-SSA to baseline methods in simulated experiments and show top balance between performance and safety across various tasks without parameter tuning.
    \item We verify p-SSA on a Unitree G1 humanoid robot in challenging safe tele-operation tasks.
\end{itemize}

The rest of the paper is organized as follows.
Section \ref{sec:related} reviews related indirect safe control approaches.
Section \ref{sec:problem} formulates the dexterous safe control problem and shows the conditions under which infeasibility arises.
Section \ref{sec:method} presents r-SSA and p-SSA to address infeasible safe control constraints.
In Section \ref{sec:experiment}, we present simulation results and hardware demonstrations of dexterous safety, followed by limitations in Section \ref{sec:limitation} and conclusions in Section \ref{sec:conclusion}.

\section{Related Works}
\label{sec:related}

In this section, we review existing works that address safety in robot control in indirect (model-based) fashion.
Indirect approaches require explicitly modeled robot dynamics and derive safe control laws according to safety specifications.


Indirect safe control backbones include safe set algorithms (SSA)  \cite{liu2014control, chen2023sis}, control barrier functions (CBFs) \cite{ames2016control, xiao2019control}, and Hamilton–Jacobi (HJ) reachability \cite{choi2021robust}.
While being different in the composition of safe control laws, all the above approaches quantifies safety using an energy function \cite{wei2019unified}.
With modeled system dynamics, optimization-based safe control laws are derived to drive the energy function below (or above if negating the sign) a critical level, satisfying Lyapunov-like conditions which guarantees invariance within a safe set.
In the past decade, indirect safe control methods have been applied to a wide range of applications such as safe quadruped navigation \cite{yun2024safe, he2024agile, molnar2021model}, safe human-robot interaction \cite{liu2023proactive, lin2017real, pandya2024multimodal, liu2022safe}, safe learning \cite{zhao2021zeroviolation}, locomotion \cite{choi2023constraint}, and high-speed drones \cite{singletary2022onboard}.
Prior work also studies more general settings with time-varying factors \cite{glotfelter2019hybrid, chen2023sia}, model uncertainty \cite{taylor2020adaptive, dawson2022safe}, and model mismatch \cite{taylor2020learning}.
All the above works only consider a single safety constraint, making the optimization for safe control normally feasible given adequate control limits.
In dexterous safety in cluttered environments, however, we often need multiple safety constraints to fully capture the safety conditions which also make the optimization sometimes infeasible.
As mentioned in Section \ref{sec:intro}, there is not existing work that handles such tasks with feasibility guarantees.
In a cluttered dynamic environment, there may also exist situations that are physically impossible for the humanoid to escape from (e.g., trapped in a crowd of people).
Prior work usually ignore these situations.
In those cases, the question is indeed not to satisfy the safety specification anymore, but how to achieve minimal violations.
With minimally relaxed control constraints, our approaches automatically fulfill that purpose as well.



\section{Dexterous Safety for Articulated Robots}\label{sec:problem}

In this section, we will formulate the problem of dexterous safety for articulated robots such as humanoids.
We first introduce preliminaries, including system modeling, safety specification, and basic indirect safe control approaches that provide theoretical guarantees.
Then, we formulate dexterous safety control where the robot interacts with the environment under multiple safety constraints.

\subsection{Safe Control Preliminaries}

\paragraph{Robot Dynamics}
We consider control-affine robot dynamics with bounded control.
Let $x\in\cX\subset \RR^{N_x}$ be the system state and $u\in\cU$ be the control input.
Let $\cU\defeq\{u\in\RR^{N_u} \mid u^- \leq u \leq u^+ \}$ where $u^-$ and $u^+$ are the lower and upper control limit respectively.
The dynamics is then given by
\begin{equation}\label{eq:dynamics}
    \dot{x} = f(x) + g(x)u, ~ u \in \cU,
\end{equation}
where $f: \RR^{N_x} \mapsto \RR^{N_x}$ and $g: \RR^{N_x} \mapsto \RR^{N_x\times N_u}$ are both locally Lipschitz continuous.

\paragraph{Constrained Robot Tasks}
Let $\cJ$ denote an arbitrary objective function to be minimized by the robot.
For instance, $\cJ$ may measure the relative distance to some goal location in navigation tasks.
The robot should also satisfy some given constraint by staying within $\cX_S$ (i.e., \textit{spec set}), a subset of the state space $\cX$.
$\cX_S$ is assumed to be the zero sublevel set of some piecewise smooth \textit{energy function} $\phi_0\defeq \cX \mapsto \RR$, i.e., $\cX_S \defeq \{x\in \cX \mid \phi_0(x) \leq 0\}$.
Both $\cX_S$ and $\phi_0$ are task specific.
For instance, $\phi_0 = d_\mathrm{min} - d$ keeps the relative distance $d$  to an obstacle above $d_\textrm{min}$, while $\phi_0=\|\hat{z}-[0, 0, 1]^\top\|_2-\epsilon$ keeps the z-axis of the robot upright up to an error of $\epsilon$.
Hence, we are interested in the following constrained task
\begin{align}\label{prob:constrained_task}
\minimizewrt{{u}}~~ & \cJ(x,u)   \\ \nonumber
\st~~ & \phi_0(x)\leq 0 \\ \nonumber
& u\in\cU 
\end{align}

\paragraph{Safe Control Backbone}

\eqref{prob:constrained_task} has been studied by a broad range of literature on indirect safe control methods such as the safe set algorithm (SSA) \cite{liu2014control} and control barrier functions (CBF) \cite{ames2014control}.
Both SSA and CBF derive safe control laws to restrict $\dot{\phi}_0$ following Lyapunov-like conditions and render a \textit{safe set} $\cX_\mathrm{safe}\subseteq\cX_\cS$ forward invariant.
Namely, if the state $x$ is already within $\cX_\mathrm{safe}$, it should never leave that set, and consequently, stay within $\cX_\cS$.
In some cases, control $u$ does not appear in $\dot\phi_0$ (e.g., $\dot\phi_0 =  - \dot d$ does not depend on the acceleration input for a second-order system), preventing the control input from driving the system to safety.
To solve that issue, the safe set algorithm (SSA) \cite{liu2014control} provides a systematic approach to design an alternative energy function $\phi$ to handle general relative degrees ($>1$) between $\phi_0$ and the control.
SSA designs a continuous and piece-wise smooth energy function $\phi \defeq \cX \mapsto \RR$ (a.k.a. the {\textit{safety index}}).
The general form of an $n^\mathrm{th}$ ($n\geq 0$) order safety index $\phi$ is given as
$\phi = (1+a_1 s)(1+a_2 s)\dots(1+a_n s)\phi_0$ where $s$ is the differentiation operator.
$\phi$ should satisfy that (a) the roots of the characteristic equation $\prod_{i=1}^n(1+a_i s) = 0$ are all negative real (to avoid overshooting of $\phi_0$), (b) $\phi_0^{(n)}$ has relative degree one to the control input.
$\phi$ is alternatively expanded to
\begin{equation}\label{def:phi_root}
    \phi \defeq \phi_0 + \textstyle\sum_{i=1}^{n}k_i \phi^{(i)}_0.
\end{equation}
where $\phi_0^{(i)}$ is the $i^\mathrm{th}$ time derivative of $\phi_0$.
Given $\phi$, SSA derives the following control constraint:
\begin{equation}\label{eq:safe_control_law}
    \dot{\phi}(x,u) \leq -\eta~\mathrm{if}~\phi(x) \geq 0 
\end{equation}
for some constant $\eta>0$.
With that, \eqref{prob:constrained_task} becomes
\begin{subequations}\label{prob:single_safe_control}
\begin{align}
\minimizewrt{{u}}~~ & \cJ(x,u)   \\
\st~~ & \dot{\phi}(x,u) \leq -\eta~\mathrm{if}~\phi(x) \geq 0 \label{eq:single_safe_control_phi_constr} \\
& u\in\cU \label{eq:single_safe_control_control_limit}
\end{align}
\end{subequations}
Solving the above optimization yields safe control $u_\mathrm{safe}$ that enforces the safety constraint, i.e.,  $\phi_0\leq 0$ \cite{liu2014control, chen2023sis}.
Notably, CBF also handles general relative degree between $\phi_0$ and the control \cite{wang2023high} and yields a similar constrained control problem to \eqref{prob:single_safe_control}.
In this paper, we focus on SSA-based approaches without loss of generality, since our contributions are in fact compatible with a family of energy-based safe controllers \cite{wei2019unified}, including but not limited to SSA and CBF.


\subsection{Dexterous Safety}


\eqref{prob:single_safe_control} provides a principled approach to enforce a single constraint $\phi_0$ under input limits $\cU$.
As discussed in Section \ref{sec:intro}, a single $\phi_0$ can rarely suffice for dexterous safety in cluttered environments.
Hence, we consider one constraint for each robot body and obstacle pair.
We will end up with a set of $M$ constraints, with $\phi_{0,i}$ being the energy function for the $i^\mathrm{th} (i\in[M])$ collision pair and $\cX_{\cS,i}$ the corresponding spec set.
Let $\phi_i$ be the corresponding $n^\mathrm{th}$ order safety index for $\phi_{0,i}$ similar to \eqref{def:phi_root}, \eqref{prob:single_safe_control}
can be naturally extended to a multi-constraint version:
\begin{subequations}\label{prob:multi_safe_control}
\begin{align}
\minimizewrt{{u}}~~ & \cJ(x,u)  \\
\st~~ & \dot{\bphi}(x,u) \leq -\bETA~\mathrm{if}~\bphi(x) \geq 0  \label{eq:multi_safe_control_phi_constr} \\ 
& u\in\cU \label{eq:multi_safe_control_contol_limit}
\end{align}
\end{subequations}
where $\bphi\defeq[\phi_1,\phi_2,\dots,\phi_M]^\top$ and $\bETA\defeq [\eta_1, \eta_2, \dots, \eta_M]^\top$, $\eta_i > 0$.
If \eqref{prob:multi_safe_control} can be solved at all times, some safe set $\cX_{\mathrm{safe},i}\subseteq\cX_{\cS,i}$ will be rendered forward invariant for each constraint $i$.
Then, all constraints are enforced and our problem is well solved.
However, for dexterous safety in cluttered environments, \eqref{prob:multi_safe_control} can easily be infeasible, as will be explained next, making our problem particularly challenging.

Without loss of generality, we assume $\cJ$ to be a quadratic objective.
Such an assumption makes the safe control problem a quadratic programming (QP) which can be efficiently solved by off-the-shelf solvers.
In the rest of this paper, we refer to safe control problems like \eqref{prob:single_safe_control} and \eqref{prob:multi_safe_control} as ``QP'' for simplicity.

\subsection{Infeasible Safe Control Problems}\label{sec:infeas_safe_control_problem}



To analyze the feasibility of multi-constraint QPs, we first see how the single-constraint QP case works.
\eqref{prob:single_safe_control} is feasible when there exists a control within $\cU$ to reduce positive $\phi$ values for all states in $\cX$.
This objective can normally be achieved by properly selecting $k_i$ coefficients in \eqref{def:phi_root} via \textit{Safety Index Synthesis} (SIS) \cite{zhao2023sos, chen2023sis}, formally given by \Cref{problem:synthesis}.
\begin{problem} [Safety Index Synthesis] \label{problem:synthesis}
    Find safety index as $\phi \defeq \phi_0 + \sum_{i=1}^{n}k_i \phi^{(i)}_0$ with parameter $\theta\in\Theta\defeq\{[k_1,k_2,\dots,k_{n}] \mid k_i\in\RR,k_i\geq 0,\forall i \}$, such that
    \begin{equation}\label{eq:synthesis}
        \forall x\in\cX ~ \st ~ \phi(x)\geq 0, \minimizewrt{u\in\cU} \dot{\phi}(x,u) < -\eta
    \end{equation}
\end{problem}
SIS is non-trivial to solve since the condition in \eqref{eq:synthesis} must hold for a continuous state space, leading to an infinite number of constraints.
In the literature, SIS has been solved via sum-of-square programming \cite{zhao2023sos, chen2023sis} which is already expensive for single-constraint cases, and normally guarantees QP feasibility only in restricted state, control, and task spaces.
Notably, the synthesis of a single feasible energy function $\phi$ primarily concerns about the compatibility between the control constraint \eqref{eq:single_safe_control_phi_constr} and the control limits \eqref{eq:single_safe_control_control_limit}, since single-constraint QPs with unbounded control are always feasible \cite{liu2014control}.
If there are multiple energy functions $\{\phi_i\}$ (more than 200 in our case), SIS needs to additionally handle potential inconsistencies within $\phi_i$'s themselves, which is extremely challenging.
SIS approaches that handle such problems are not found in the literature yet.

\begin{figure}[htbp]
    \centering
    \begin{subfigure}[b]{0.3\linewidth}
        \centering
        \includegraphics[width=\linewidth]{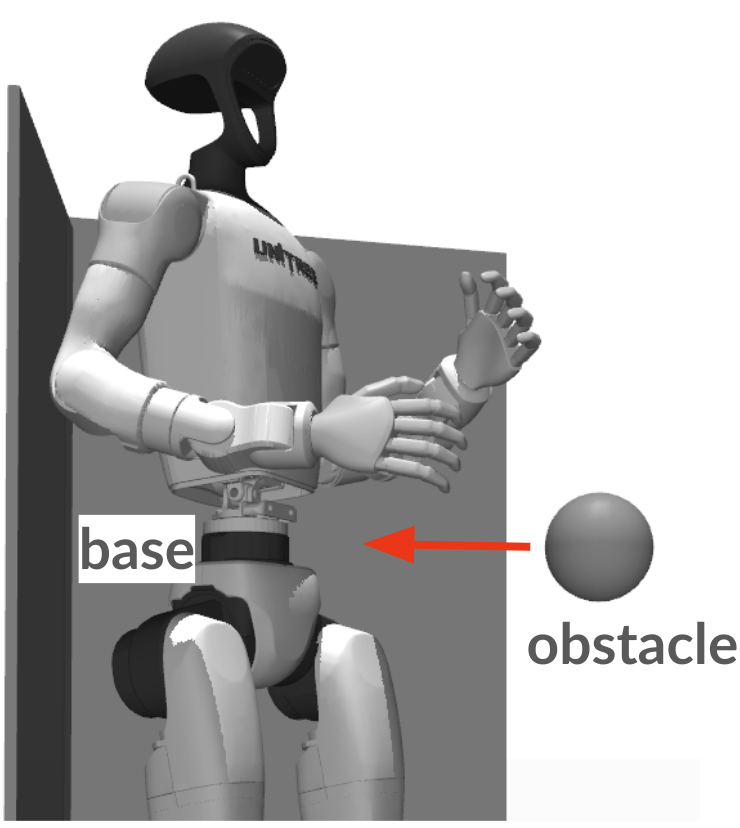}
        \caption{Inherent Infeasibility}
        \label{fig:infeas_1}
    \end{subfigure}
    \hfill
    \begin{subfigure}[b]{0.3\linewidth}
        \centering
        \includegraphics[width=\linewidth]{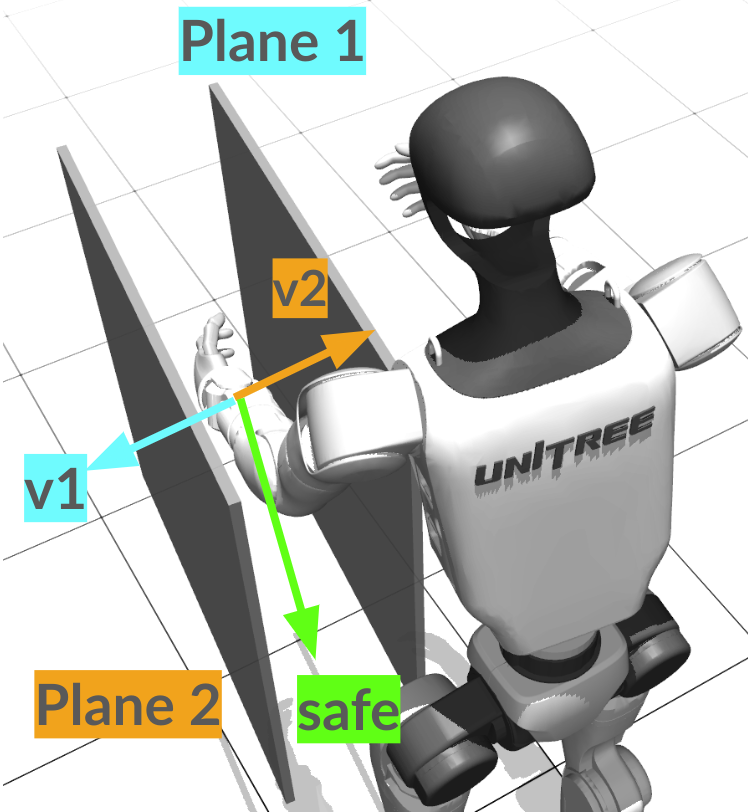}
        \caption{Method Infeasibility}
        \label{fig:infeas_2}
    \end{subfigure}
    \hfill
    \begin{subfigure}[b]{0.32\linewidth}
        \centering
        \includegraphics[width=\linewidth]{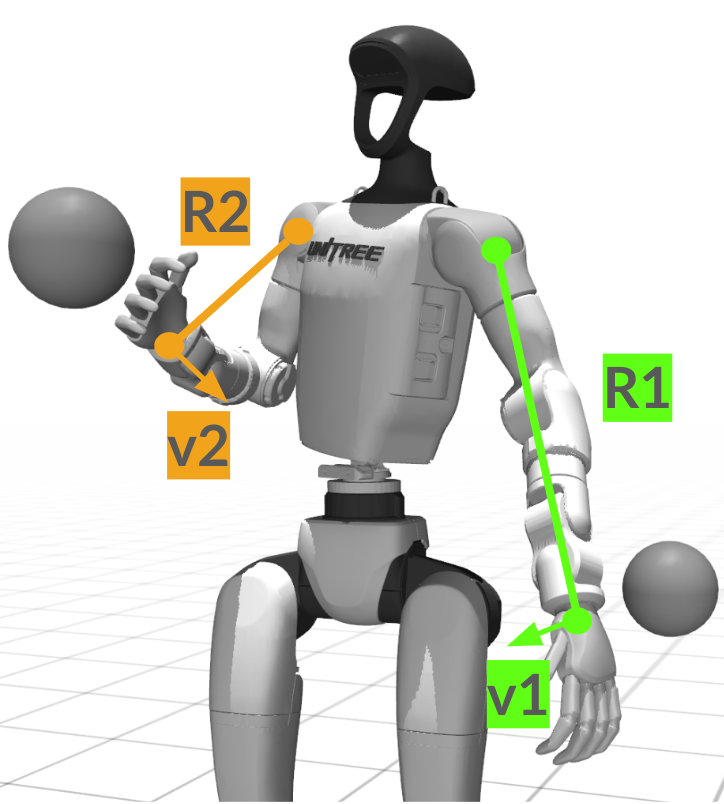}
        \caption{Kinematics Infeasibility}
        \label{fig:infeas_3}
    \end{subfigure}
    \caption{Possible scenarios where \eqref{prob:multi_safe_control} can be infeasible. The humanoid should avoid collision with all obstacles (e.g., planes and spheres) in gray.}
    \label{fig:infeas_analysis}
\end{figure}

In this paper, we argue that, it is intractable to make \eqref{prob:multi_safe_control} or any similar formulations always feasible for dexterous safety in cluttered environments under actuation bounds.
To help illustrations, we explain possible infeasibility of three types: inherent infeasibility, method infeasibility, and kinematics infeasibility (see \cref{fig:infeas_analysis}).

\paragraph{Inherent Infeasibility}
In cluttered environments with dynamic obstacles, there can be cases where collisions are physically inevitable.
For instance, when an obstacle moves towards the base of a humanoid that is already unable to move due to other constraints (see \cref{fig:infeas_1}), there is no control that can prevent collisions, making \eqref{prob:multi_safe_control} infeasible.
Such inherent infeasibility can hardly be avoided without restricting the operation conditions of humanoids.

\paragraph{Method Infeasibility}
Since each $\phi_i$ in \eqref{eq:multi_safe_control_phi_constr} concerns a robot body and obstacle pair, different $\phi_i$'s may generate conflicts in some cases.
Consider the collision avoidance between the humanoid wrist and two planar obstacles.
One can construct an energy function for each wrist-obstacle pair, e.g., $\phi_i = d_\mathrm{min} - d_i$ where $d_i$ is the distance from the surface of the wrist to plane $i$ for $i=1,2$ .
According to \eqref{eq:multi_safe_control_phi_constr}, the corresponding control constraints will be $\dot{\phi}_i = -\dot{d}_i = -v_i \leq -\eta_i \Rightarrow v_i \geq \eta_i$ which simply constrains the wrist to move away from the plane with at least $\eta_i$ velocity ($\eta_i>0$) when $d_i\leq d_\mathrm{min}$.
Such design is mostly effective assuming accurate velocity tracking.
However, in some cases, even those two control constraints may not be compatible.
For example, when the humanoid operates between two planes parallel to each other (e.g., when manipulating objects in shelves) with a total distance to both sides $d_1+d_2$ less than $d_\mathrm{min}$, both $\phi_1$  and $\phi_2$ are non-negative.
Then, each $\phi_i$ would require the wrist to move towards an opposite direction, i.e., $v_1 \geq \eta_1 > 0$ and $v_2 \geq \eta_2 > 0$, which is impossible and renders \eqref{prob:multi_safe_control} infeasible (see \cref{fig:infeas_2}).
However, the humanoid can simply move the arm parallel to the planes until exiting from the openings to be safe (green direction in \cref{fig:infeas_2}).
Hence, different from the previous case, a safe control is not impossible in this case, but rather not found by the QP \eqref{prob:multi_safe_control} with the designed $\phi_i$.
We refer to such failures as method infeasibility.
Although the QP in certain cases (e.g., the example above) can be made feasible by re-designing the control law, it hardly generalizes since we cannot know all possible obstacle configurations beforehand in cluttered environments.

\paragraph{Kinematics Infeasibility}

Infeasibility can also be caused by the complex kinematics chain of humanoids.
Consider a humanoid avoiding hand collision with the sphere nearby in \cref{fig:infeas_3}.
With a similar definition of $\phi_i$ to the above, the control constraints \eqref{eq:multi_safe_control_phi_constr} would require the hand $i$ to move away with a velocity of $v_i \geq \eta_i$ for $i=1,2$ when being too close to the obstacle.
Restricted by the kinematics capability, the maximum velocity at which the hand can move depends on the actual joint configurations.
In \cref{fig:infeas_3}, with a more extended arm pose, the left hand would have a higher velocity limit than the right hand.
Hence, a relatively large $\eta_i$ may work for the left hand but make the QP infeasible when handling the right hand.
While a universally small $\eta_i$ may improve QP feasibility, that would make the robot overly insensitive to potential collisions.
Designing energy functions $\phi_i$ to be compatible with each body on the humanoid in general cases remains a challenge.

The environment configurations shown in \cref{fig:infeas_analysis} are merely toy examples while still making \eqref{prob:multi_safe_control} infeasible.
To address dexterous safety in cluttered environments, we will have significantly more constraints to handle the complex obstacle configurations.
Several scenarios like those in \cref{fig:infeas_analysis} may even be coupling, leading to QP infeasibility that can hardly be mitigated in practice.
In that regard, we do not aim to construct persistently feasible QPs but instead desire a practical method that minimizes violations of control constraints when the QP becomes infeasible.

\section{Projected Safe Set Algorithm for Infeasible Constraint Sets}
\label{sec:method}

In this section, we propose several approaches to handling infeasible safe control problems.
To facilitate further derivation, we first expand the safe control constraints in \eqref{prob:multi_safe_control} by plugging in the system dynamics as follows.
\begin{align}\label{eq:expand_safe_constraint}
\dot{\bphi}(x,u) &= \underbrace{\frac{\partial\bphi}{\partial x}f(x)}_{L_f\bphi(x)} + \underbrace{\frac{\partial\bphi}{\partial x}g(x)}_{L_g\bphi(x)} u \\\nonumber
&= L_f\bphi(x) + L_g\bphi(x) u \leq -\bETA
\end{align}
While our approaches are compatible with arbitrary task objective $\cJ$, we instantiate an example in this section to aid discussions.
We assume that \eqref{prob:multi_safe_control} operates as a safety filter at the downstream of some nominal controller.
With nominal control signal $u_\mathrm{ref}$, we can set $\cJ(x,u) = \|u-u_\mathrm{ref}\|_{2,Q}^2$ to compute a minimally invasive control $u_\mathrm{safe}$ that satisfies safety constraints.
Incorporating the control limits as well, \eqref{prob:multi_safe_control} can be written as
\begin{subequations}\label{prob:naive_ssa}
\begin{align}
\minimizewrt{{u}}~~ & \|u-u_\mathrm{ref}\|_{2,Q}^2   \\
\st~~ & L_f\bphi(x) + L_g\bphi(x) u \leq -\bETA~\mathrm{if}~\bphi(x) \geq 0  \label{eq:infeas_ssa_contr_lie} \\ 
& u^- \leq u \leq u^+ \label{eq:control_limit}
\end{align}
\end{subequations}

\subsection{Relaxed Safe Set Algorithm}\label{sec:rssa}

When \eqref{prob:naive_ssa} becomes infeasible due to complex humanoid-environment interactions, the most straight-forward remedy is to incorporate slack variables that relaxes the constraints to allow feasible solutions.
Specifically, we introduce positive slack variables only for each of the safety constraints in \eqref{eq:infeas_ssa_contr_lie} since the control limits cannot be relaxed.
Hence, we have
\begin{subequations}\label{prob:rssa}
\begin{align}
\minimizewrt{{u, \bs}}~~ & \|u-u_\mathrm{ref}\|_{2,Q}^2+ \frac{1}{p}(\|\bs\|_{p,Q^{rssa}_s})^p  \\
\st~~ & L_f\bphi(x) + L_g\bphi(x) u \leq -\bETA + \bs ~\mathrm{if}~\bphi(x) \geq 0 \label{eq:safe_contr_rssa}\\ 
& u^- \leq u \leq u^+ \\
&  \bs \geq 0 
\end{align}
\end{subequations}
where $\bs\in\RR^{M}$ is the slack variable.
We regularize $\bs$ measured in $Q^{rssa}_s$-weighted $p-$norm, given by
\begin{equation}
\|\bs\|_{p,Q^{rssa}_s} = \left(\sum_{i=1}^{M}Q^{rssa}_{s,i}|\bs_i|^p\right)^{1/p}
\end{equation}
where $Q^{rssa}_s$ is a diagonal matrix with positive coefficients.
We refer to \eqref{prob:rssa} as Relaxed Safe Set Algorithm (r-SSA).
The solved $\bs$ indicates the cost of safety violations, and should be as close to zero as possible to try to respect the safety constraints.
As will be shown later, r-SSA can effectively produce safe control when the naive SSA \eqref{prob:naive_ssa} becomes infeasible, and enhances humanoid safety in cluttered environments.

Importantly, r-SSA optimizes a combination of both performance objective (i.e., the first term) and safety objectives (i.e., the second term), which may be conflicting in general.
It also balances multiple safety objectives represented by each $\phi_i$.
r-SSA may prioritize the performance (i.e., reference tracking) and accept large slack variables (i.e., significant safety violations) if partial safety objectives are dominated, especially when the weighting parameters $Q$ and $Q^{rssa}_s$ are not properly tuned.
As a result, r-SSA may still lead to critical safety failures in practice unless specifically tuned for each task.
This challenge motivates us to take another step by removing potential racing conditions between the two objectives, which will be covered in the next section.

\subsection{Projected Safe Set Algorithm}\label{sec:pssa}



In this section, we propose the Projected Safe Set Algorithm (p-SSA) that improves over r-SSA by always respecting the safety constraints to the maximal extend.
The core idea behind p-SSA is to first project the current safe control constraint set, which can be infeasible, to the nearest feasible set.
The $u_\mathrm{safe}$ is only solved with the projected constraint set which is guaranteed to be feasible by the projection operation.
It can be shown that with such decoupling, p-SSA always operates within the maximal feasible region indicated by the given constraint set, while being totally tuning-free.

In phase I, the p-SSA first resolves infeasible safe control constraints in \eqref{prob:naive_ssa} by projecting the constraint set \eqref{eq:infeas_ssa_contr_lie}
 and \eqref{eq:control_limit} on to the nearest feasible region, measured by the $p$-norm of  total relaxation.
Specifically, we solve the following optimization
\begin{subequations}\label{prob:pssa_phase_1}
\begin{align}
\minimizewrt{\bs}~~ & \frac{1}{p}(\|\bs\|_{p,Q^{pssa}_s})^p  \\
\st~~ & L_f\bphi(x) + L_g\bphi(x) u \leq -\bETA + \bs ~\mathrm{if}~\bphi(x) \geq 0 \label{eq:safe_contr_pssa_phase_1}\\ 
& u^- \leq u \leq u^+ \label{eq:control_limit_pssa_phase_1}\\
&  \bs \geq 0
\end{align}
\end{subequations}
to compute an optimal slack variable $\bs^*$.
Then, in phase II, we solve $u_\mathrm{safe}$ with the solved relaxation $\bs^*$
\begin{subequations}\label{prob:pssa_phase_2}
\begin{align}
\minimizewrt{u}~~ & \|u-u_\mathrm{ref}\|_{2,Q}^2  \\
\st~~ & L_f\bphi(x) + L_g\bphi(x) u \leq -\bETA + \bs^* ~\mathrm{if}~\bphi(x) \geq 0 \label{eq:safe_contr_pssa_phase_2}\\ 
& u^- \leq u \leq u^+ \label{eq:control_limit_pssa_phase_2}
\end{align}
\end{subequations}
Since $\bs^*$ is feasible for \eqref{prob:pssa_phase_1}, we know that the constraints \eqref{eq:safe_contr_pssa_phase_1} with \eqref{eq:control_limit_pssa_phase_1} will be made feasible if relaxed by $\bs^*$ (i.e.,  \eqref{eq:safe_contr_pssa_phase_2} and \eqref{eq:control_limit_pssa_phase_2} ).
Hence, phase II is guaranteed to be feasible without additional relaxation. 
Notably, p-SSA does not involve any direct trade-off between performance and safety since they are optimized independently via \eqref{prob:pssa_phase_1} and \eqref{prob:pssa_phase_2}.
Meanwhile, p-SSA guarantees to operate with minimal safety violations thanks to Phase I.
As will be shown in the next section, p-SSA, in fact, achieves the top performance across various task settings without parameter tuning, while r-SSA only matches p-SSA performance with careful tuning.

\textbf{Remark.} Unlike $Q^{rssa}_s$, $Q^{pssa}_s$ only influences the balance among multiple constraints.
In practice, $Q^{pssa}_s$ should be chosen to reflect the relative importance of different constraints.
For example, more weight can be assigned to a constraint if it covers a critical aspect such as the safety of a high-torque link, or if it is closer to being violated (e.g., $\phi_i$ closer to zero).
In this paper, we set $Q^{pssa}_s$ to identity and leave the investigation of smart ways to balance multiple safety constraints for future work.

When implementing the above approaches, one needs to derive the two Lie derivatives, $L_f\bphi(x)$ and $L_g\bphi(x)$, based on the designed energy functions $\phi_i$ and plug into \eqref{eq:safe_contr_rssa} for r-SSA or \eqref{eq:safe_contr_pssa_phase_1} and \eqref{eq:safe_contr_pssa_phase_2} for p-SSA to complete the control constraints.
For readability, we provide an example derivation of control constraints based on first-order energy functions designed for both robot-obstacle collision and self-collision in Appendix \ref{append:safe_control_constraint}.

\section{Experiments}
\label{sec:experiment}

In this section, we aim to showcase how our proposed methods handle infeasible safe control problems and ultimately improve safety for humanoids in cluttered environments.
Through both simulation and hardware experiments, we will answer the following key questions.

\begin{itemize}
\item \textbf{Q1}: How often does \eqref{prob:naive_ssa} become infeasible in dexterous safety tasks in cluttered environments?
\item \textbf{Q2}: Does r-SSA and p-SSA mitigate infeasible QPs and improve humanoid safety in cluttered environments?
\item \textbf{Q3}: How does r-SSA compare to p-SSA in terms of parameter tuning and performance?
\item \textbf{Q4}: How does p-SSA perform in real-world dexterous safety tasks?
\end{itemize}

\begin{figure}[ht]
    \centering
    \includegraphics[width=1.0\linewidth]{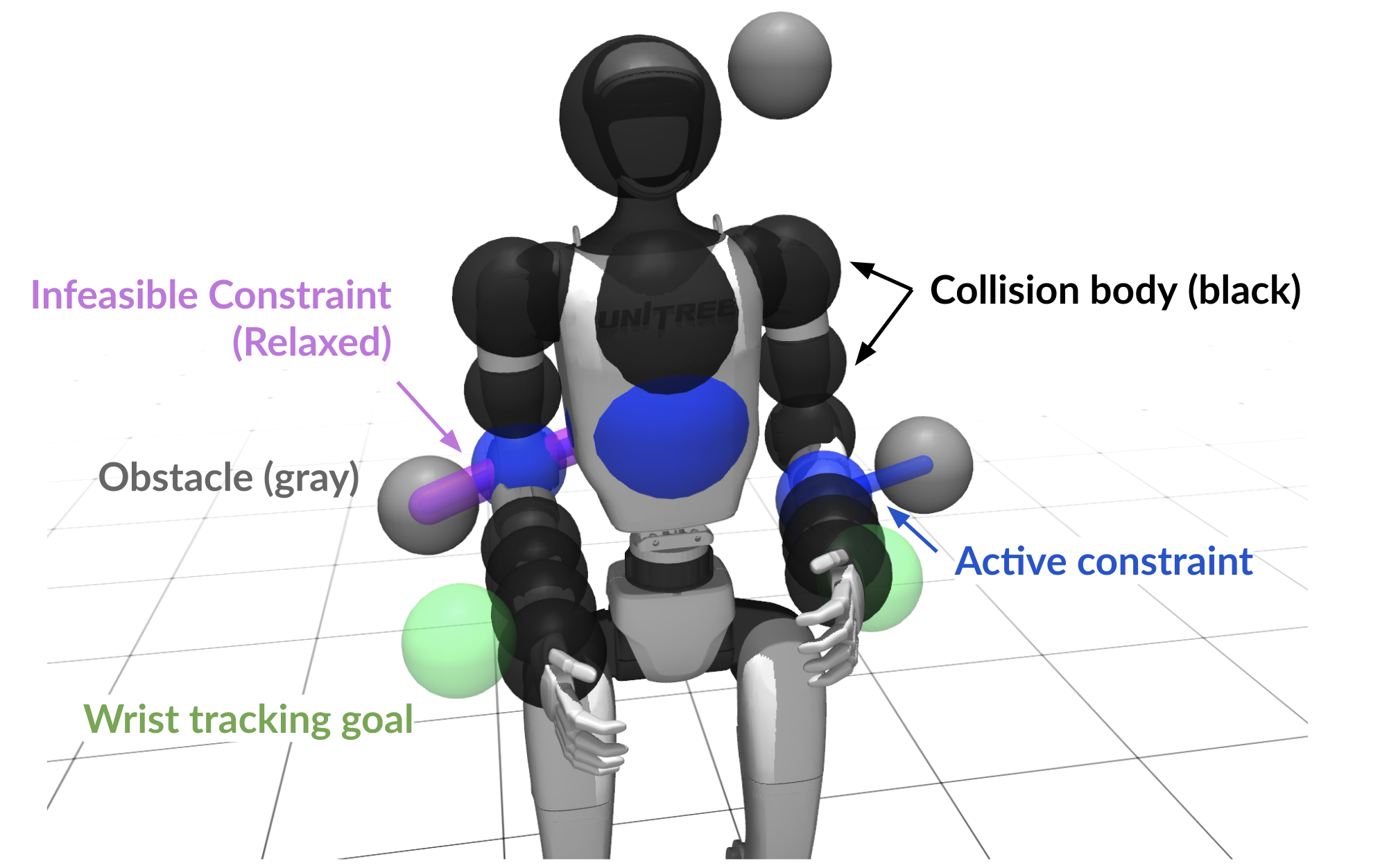}
    \caption{Unitree G1 humanoid robot in MuJoCo simulation performing safe wrist location tracking.
    The humanoid tracks the goal (green) with its wrist while preventing collisions between robot bodies (black) and obstacles (gray) and self-collision.
    There are three active control constraints (blue) triggered by collision bodies being too close, while two of them are infeasible and relaxed by p-SSA (purple).
    The infeasibility is caused by the right arm trying to avoid both the obstacle and the torso at the same time.
    }
    \label{fig:system}
\end{figure}

\subsection{Robot Modeling}\label{sec:robot_model}

We consider the Unitree G1 humanoid robot in both MuJoCo simulation and real setup.
We model the G1 dynamics with two variants to represent different use cases.

\textbf{G1FixedBase}:
This configuration considers only the upper body joints while keeping the base fixed.
This model is useful for tasks where the humanoid is expected to perform tasks in-place, such as organizing objects on a fixed shelf.
This model features \textbf{17 DoFs}—7 DoFs for each arm and 3 for the waist—with the pelvis fixed to the world frame.

\textbf{G1WholeBody}:
This configuration includes \textbf{20 DoFs}, consisting of 17 DoFs from G1FixedBase and 3 for base motion, modeled as a floating base.
It is designed to assess safe control for humanoid robots as mobile manipulators, but isolating the challenges induced by locomotion.

With either configuration, we model the humanoid dynamics using a first order integrator model, i.e., $\dot{x} = f(x) + g(x)u = u$ where $f(x) = 0$ and $g(x) = I$.
The state $x$ contains all DoF positions and $u$ the DoF velocities, assuming an accurate velocity tracker at the downstream.
Such setting simplifies the derivations while enabling us to focus on QP infeasibility.
In practice, any control-affine systems\footnote{Non-control-affine systems can still be made control-affine via dynamics extension.} in the form of \eqref{eq:dynamics} can be used.
Collision bodies on the humanoid are modeled in spheres as shown in \cref{fig:system}.
\begin{table}[htbp]
\centering
\captionsetup{width=0.95\textwidth}
\caption{Number of Bodies, Obstacles and Constraints in Test Cases}
\label{tab:test_case_size}

\begin{tabular}{lccccc}
\toprule
\textbf{Test Case} & \textbf{DoFs} & \textbf{Bodies} & \textbf{Obs} & \textbf{Self} & \textbf{Body-Obs} \\
\midrule
G1WholeBody\_SO\_V0 & 20 & 19 & 50 & 29 & 950 \\
G1WholeBody\_SO\_V1 & 20 & 19 & 10 & 29 & 190 \\
G1WholeBody\_DO\_V0 & 20 & 19 & 50 & 29 & 950\\
G1WholeBody\_DO\_V1 & 20 & 19 & 10 & 29 & 190 \\
G1FixedBase\_SO\_V0 & 17 & 19 & 10 & 29 & 190\\
G1FixedBase\_SO\_V1 & 17 & 19 & 5 & 29 & 95 \\
G1FixedBase\_DO\_V0 & 17 & 19 & 10 & 29 & 190\\
G1FixedBase\_DO\_V1 & 17 & 19 & 5 & 29 & 95\\
\bottomrule
\end{tabular}

\end{table}

\begin{figure*}[ht]
\centering
    \includegraphics[width=1.0\textwidth]{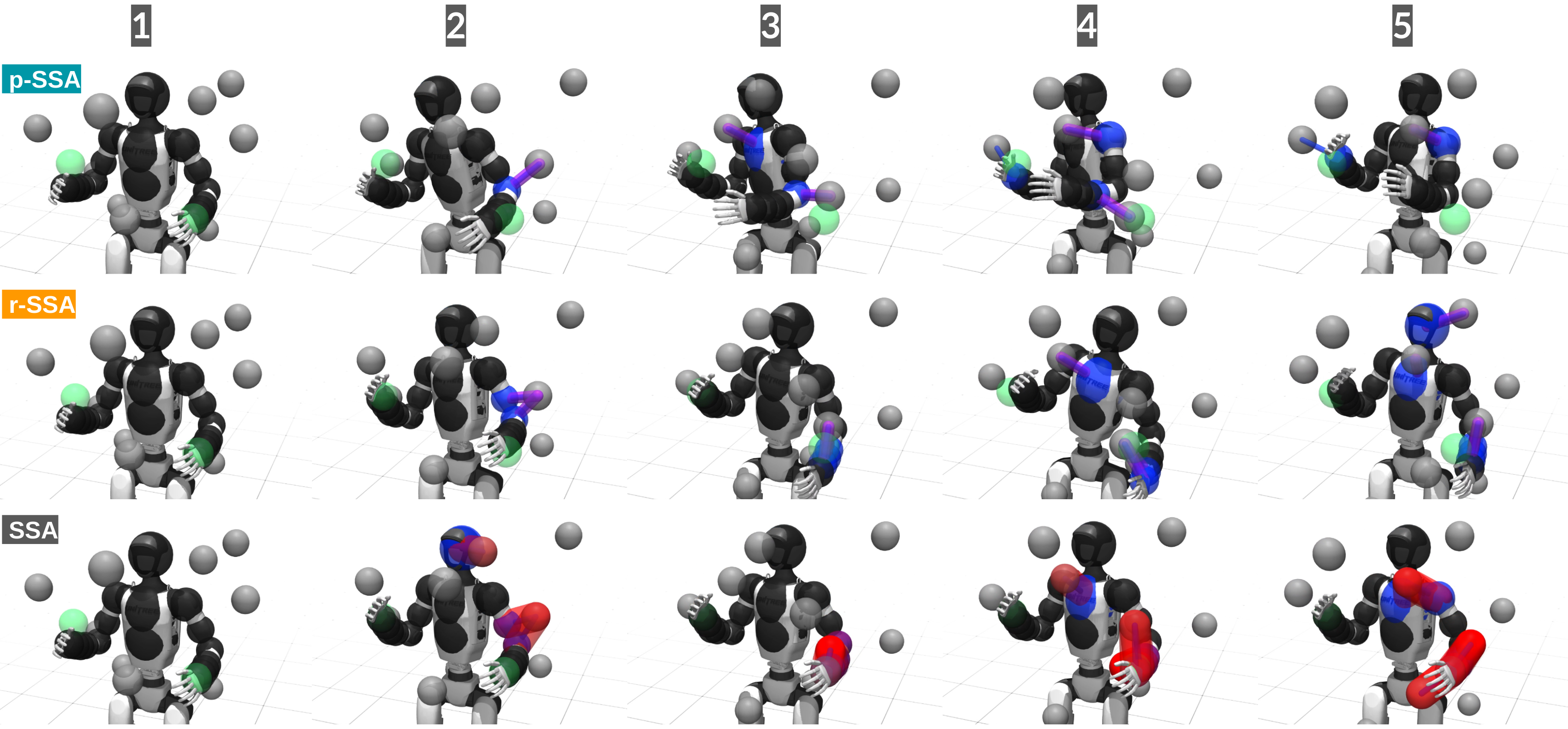}
    \caption{Comparison of safe control methods in G1FixedBase\_DO\_v0 task. Spheres and lines follow the convention in \cref{fig:system}. When an obstacle moves near the left elbow (frame 2), the QP becomes infeasible. In that case, p-SSA (top) generates control to minimize violation to control constraints, resulting in less violation (purple connection) than r-SSA (middle). Naive SSA (bottom) does not handle infeasible control constraints (thick red connection), leading to collisions (red spheres).}
\label{fig: sim_example}
\end{figure*}

\subsection{Experiment Setting}\label{sec:exp_setting}

To evaluate safe control algorithms in various cluttered environments, we design a set of dexterous safety tasks with various obstacle configurations and densities.
In all tasks, the robot is tasked to track fixed 3D goal positions for its wrists and base (if with \textbf{G1WholeBody}) while keeping at least $d_\mathrm{min,env}=0.05 m$ from external obstacles and at least $d_\mathrm{min,self}=0.01m$ for self-collision.
A new goal will spawn once the pervious one is reached.

For constraint configurations, we consider two types:
\begin{itemize}
\item \textbf{Static Obstacle (SO):} Stationary 3D spheres with $0.05m$ radius distributed uniformly within a cubic task space of size $0.8m$.
\item \textbf{Dynamic Obstacle (DO):} Moving 3D spheres that follow Brownian motion after spawning uniformly within a task space of size $2m\times 2m\times 0.8m$.
\end{itemize}

For each constraint type (SO and DO), we assign two levels of difficulty (V0 and V1) indicated by the number of obstacles.
Considering two variants of dynamics configuration (G1FixedBase and G1WholeBody), we end up with eight different dexterous safety tasks.
\Cref{tab:test_case_size} shows the number of all considered robot body-obstacle pairs as well as self-collision pairs in each task.
When counting self-collision pairs, adjacent robot bodies that are always within $d_\mathrm{min,self}$ distance are ignored.
See Appendix \ref{append:self_collision} for detailed configurations of the collision pairs.
Both the obstacles and goals are represented as 3D spheres (see \cref{fig:system}).

\subsection{Safe Control Methods}

To model safety after \eqref{prob:multi_safe_control}, we design an $0^\mathrm{th}$ order safety index $\phi^\mathrm{env}_i=\phi^\mathrm{env}_{0,i}=d_\mathrm{min,env}-d_i$ (since our dynamics is $1^\mathrm{st}$ order) for the $i^\mathrm{th}$ body-obstacle pair where $d_i$ is the body-obstacle distance.
Likewise, $\phi^\mathrm{self}_i = \phi^\mathrm{self}_{0,i} = d_\mathrm{min,self}-d_i$  covers self-collision.
The final safety index is $\bphi\defeq[\phi^\mathrm{env}_1,\dots,\phi^\mathrm{env}_{M_\mathrm{env}},\phi^\mathrm{self}_1,\dots,\phi^\mathrm{self}_{M_\mathrm{self}}]\in\RR^M$.
$\bETA$ is set to $0.5$ for each $\phi_i$ for comparison under the same sensitivity to potential collisions.
Under the same basic safe control law (i.e., $\dot{\bphi}\leq -\bETA$ when $\bphi \geq 0$), we are interested in how the following strategies handle infeasible QPs.

\begin{itemize}
\item \textbf{$p-SSA_2$}: The Projected Safe Set Algorithm introduced in \cref{sec:pssa} with $p=2$ in \eqref{prob:pssa_phase_1}. $Q=Q^{pssa}_s=I$.

\item \textbf{$r-SSA_2$}: The Relaxed Safe Set Algorithm introduced in \cref{sec:rssa} with $p=2$ in \eqref{prob:rssa}. $Q=I$. With a fixed $Q$, $Q^{rssa}_s$ decides the relative importance between performance and safety. $Q^{rssa}_s$ will be tuned in the ablation study.

\item \textbf{$SSA$}: The naive safe set algorithm given by \eqref{prob:naive_ssa}. $Q=I$. When \eqref{prob:naive_ssa} becomes infeasible, $u_\mathrm{ref}$ is directly passed to the robot since there is no special handling of infeasibility.
\end{itemize}

We use a PID policy to generate $u_\mathrm{ref}$ for goal tracking without considering safety.
Each of the above safe control methods finds a control $u$ as close to $u_\mathrm{ref}$ as possible to enforce safety constraints.
See Appendix \ref{append:safe_control_constraint} for the derivation of control constraints in \eqref{eq:safe_contr_rssa}, \eqref{eq:safe_contr_pssa_phase_1}, and \eqref{eq:safe_contr_pssa_phase_2}.



\subsection{Evaluation Metrics}

We evaluate each method in terms of both the task performance (e.g., goal tracking) and safety using a combination of metrics.
For a trajectory of length $T$:
\begin{itemize}
    \item $J$: goal tracking performance given by
        \begin{equation}
            J = \frac{1}{2T}\sum_{t,i}\exp(-d_{i,t}^2 / 0.002)
        \end{equation}
        where $d_{i,t}$ is the distance of the wrist or base $i$ to its goal at each $t$.
        
    \item $C$: control constraint satisfaction score, given by
        \begin{equation}
            C=\frac{\sum_{t}\II(s_t>0)\exp\left(-s_t^2 / 0.2\right)}{\sum_{t}\II(s_t>0)}
        \end{equation}
        where $s_t=\sum_{i}s_{i,t}$. The parameter $s_{i,t}$ is the amount of violation to the $i^\mathrm{th}$ control constraint at time $t$ and $s_t$ is the total violation at time $t$. For p-SSA, $s_{i,t}$ is the solution to \eqref{prob:pssa_phase_1}. For r-SSA, $s_{i,t}$ is solved in \eqref{prob:rssa} with the control. For SSA, $s_{i,t}=\max(\dot{\phi}_{i,t}(x_t,u^*_t)+\eta_{i}, 0)$ where $u^*_t$ is the solution to \eqref{prob:naive_ssa} if feasible, or $u_\mathrm{ref,t}$ otherwise.
        $C$ only evaluates time steps when the QP is infeasible.
        
    \item $S$: safety constraint satisfaction score, given by
        \begin{equation}
            S = \frac{\sum_{t,i}\II(d_{i,t} < d_\mathrm{min})\exp(-(d_{i,t}-d_\mathrm{min})^2 / 0.0002)}{\sum_{t,i}\II(d_{i,t} < d_\mathrm{min})}
        \end{equation}
        $S$ evaluates the average amount of safety margin $d_\mathrm{min}$ that is invaded (e.g., negative $d_{i,t}-d_\mathrm{min}$).
        $S=0$ for no violation.
        Notably, $S$ evaluates direct violation to the safety specification (e.g., collisions) as a result of violation to control constraints which is evaluted by $C$.
        Hence, they are positively related while being different metrics.

    \item $R_\mathrm{Feas}$: empirical probability of the QP to be feasible, given by
        \begin{equation}
            R_\mathrm{Feas} = 1 - \frac{1}{T}\sum_{t}\II(s_t>0)
        \end{equation}
        
\end{itemize}

\begin{figure*}[ht]
\centering
    \includegraphics[width=1.0\textwidth]{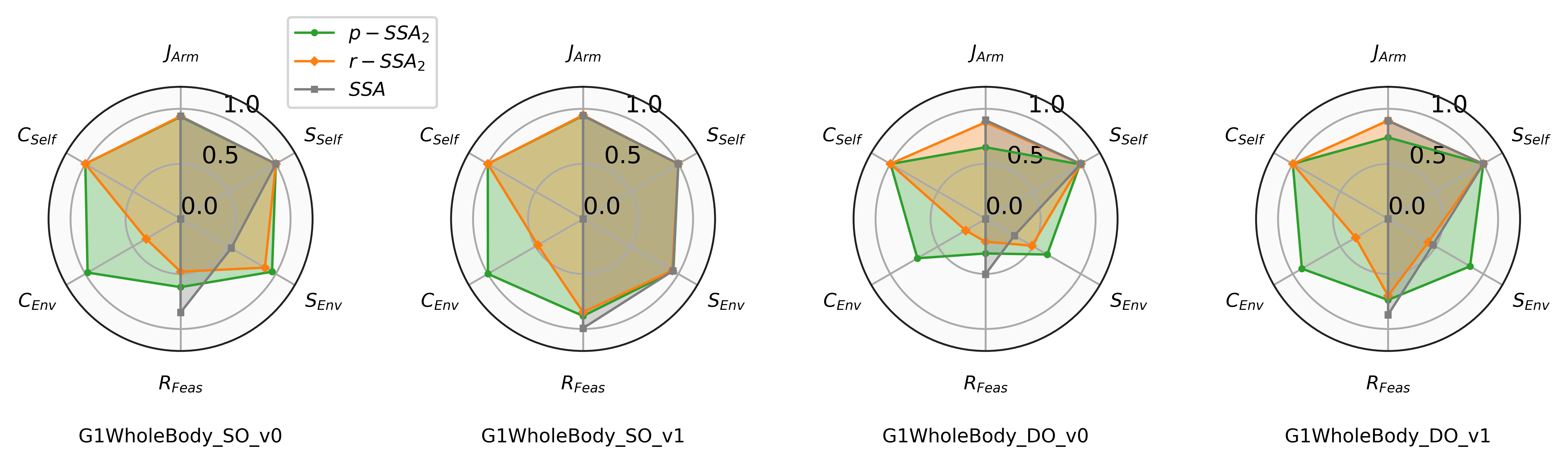}
    \caption{Performance comparison under \textbf{G1WholeBody} configuration.}
\label{fig: whole_body}
\end{figure*}
\begin{figure*}[ht]
\centering
    \includegraphics[width=1.0\textwidth]{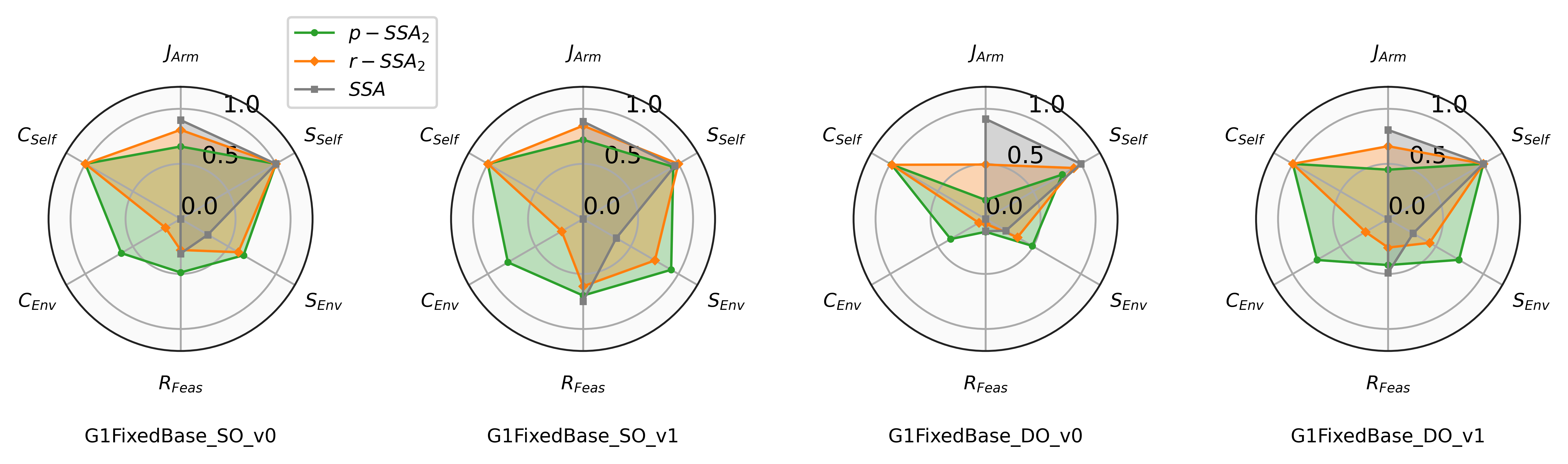}
    \caption{Performance comparison under \textbf{G1FixedBase} configuration.}
\label{fig: fixed_base}
\end{figure*}

\subsection{Overall Performance}



We apply $p-SSA_2$, $r-SSA_2$ ($Q^{rssa}_s=10I$) and $SSA$ to the eight test cases, each for 2000 steps.
We evaluate the tracking performance for arms $J_\mathrm{Arm}$, $C$ score and $S$ score computed independently for environment collision ($C_\mathrm{env}/S_\mathrm{env}$) and self-collision ($C_\mathrm{self}/S_\mathrm{self}$), and QP feasibility rate $R_\mathrm{Feas}$.
\Cref{fig: whole_body} and \Cref{fig: fixed_base} report the overall comparison.
See \cref{fig: sim_example} for examples of safety behaviors driven by the three methods.
See Appendix \ref{append:phi_compare} for the corresponding plot of (a) $\phi$ values for the left hand, (b) joint positions of the left elbow, and (c) joint positions of the left shoulder joint that mainly drive those safety behaviors.

\paragraph{QP Feasibility Rate}
We first focus on $R_\mathrm{Feas}$ (bottom of each plot).
Answering \textbf{Q1}, the QP is more likely to be infeasible with more obstacles (V0).
Comparing the robot modeling, the FixedBase variant also consistently makes QP easier to be infeasible due to the lack of mobility.
Finally, obstacle movements (DO) reduce feasibility score since the safe control computed at the current step might be made unsafe by dynamic obstacles.
Note that the naive SSA has higher $R_\mathrm{Feas}$ than other methods in many cases.
This is because naive SSA essentially ignores the obstacles if the QP is infeasible.
Since we disable the physical collision for simulated obstacles, the humanoid bodies can quickly swing through the obstacles and enter empty spaces.
That makes QPs solved by naive SSA to have less active constraints and easier to be feasible.
On the contrary, r-SSA and p-SSA normally enable the bodies to stay close to obstacles without collisions.
That keeps the number of active constraints high in the QPs to solve, making more QPs to be infeasible.

\paragraph{Minimizing Violations}

Since naive SSA directly passes the reference control upon infeasible QP, violations to control constraints are significant, resulting in negligible $C$ scores. 
Answering \textbf{Q2}, r-SSA and p-SSA significantly improve $C$ scores over naive SSA, meaning that they consistently reduce violations to control constraints when the QP is infeasible.
Comparing r-SSA to p-SSA, we see that p-SSA further out-performs r-SSA in terms of minimizing control constraint violations ($C$), and consequently minimizing safety constraint violations ($S$).
That is because p-SSA independently minimizes constraint violations, while r-SSA sometimes trades safety for performance (e.g., higher $J_\mathrm{Arm}$ in many cases).
In the next section, we perform ablation study to systematically investigate this issue.


\begin{figure*}[htbp]
    \centering
    
    \begin{subfigure}[b]{0.24\textwidth}
        \includegraphics[width=\textwidth]{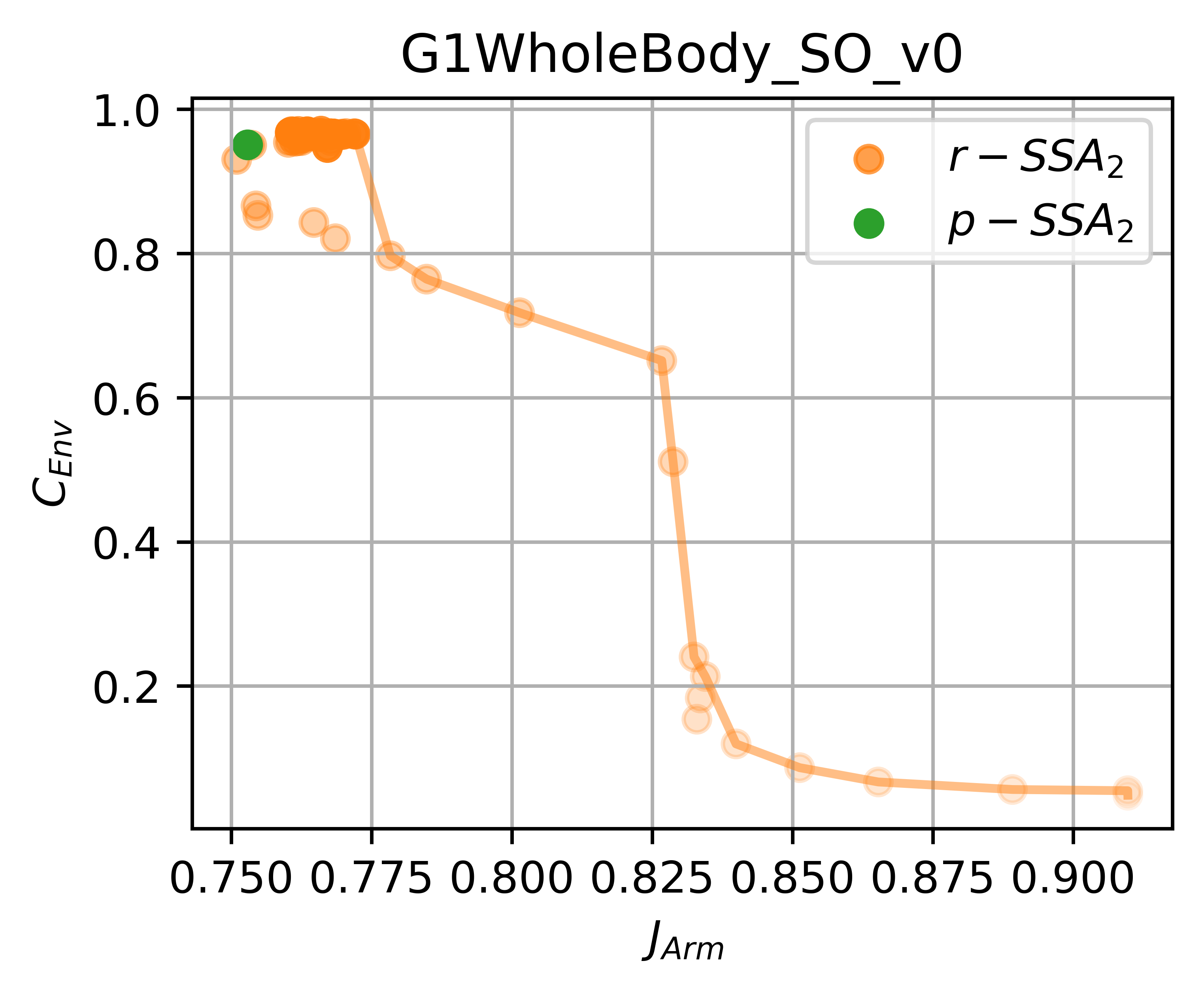}
        \caption{G1WholeBody Static Obstalce V0}
        \label{fig:pareto_G1WholeBody_SG_SO_v0}
    \end{subfigure}
    \begin{subfigure}[b]{0.24\textwidth}
        \includegraphics[width=\textwidth]{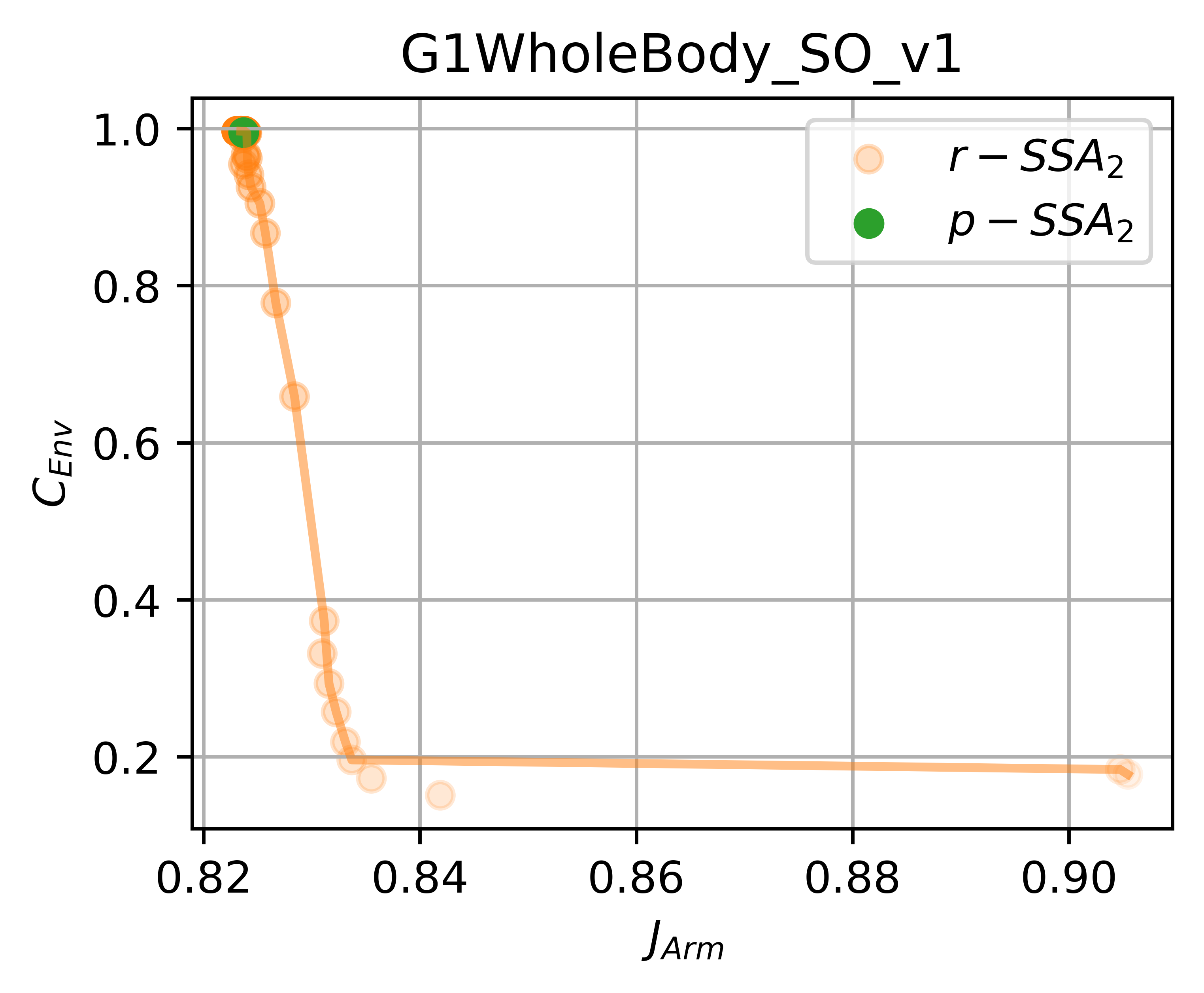}
        \caption{G1WholeBody Static Obstalce V1}
        \label{fig:pareto_G1WholeBody_SG_SO_v1}
    \end{subfigure}
    \begin{subfigure}[b]{0.24\textwidth}
        \includegraphics[width=\textwidth]{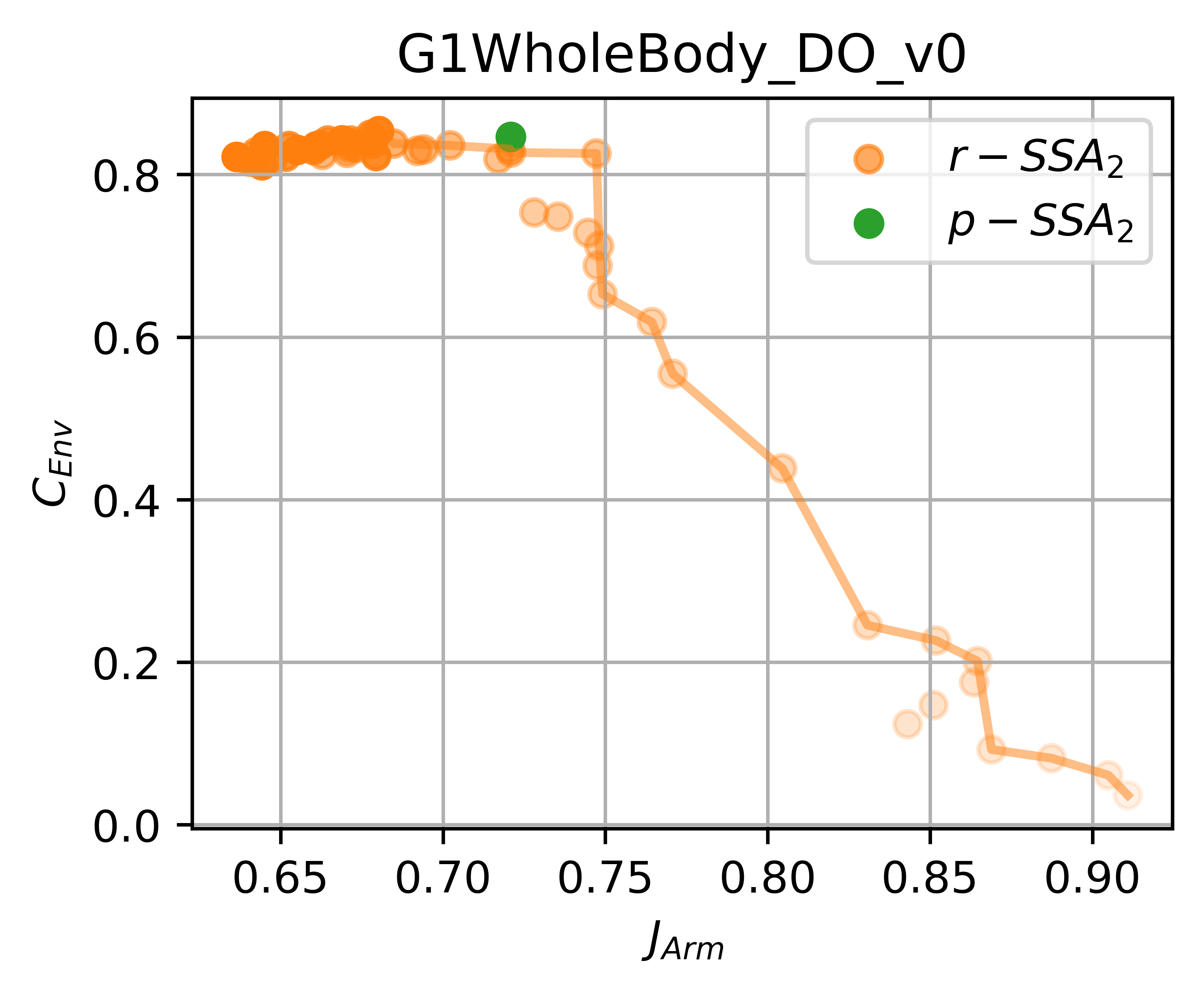}
        \caption{G1WholeBody Dynamic Obstalce V0}
        \label{fig:pareto_G1WholeBody_SG_DO_v0}
    \end{subfigure}
    \begin{subfigure}[b]{0.24\textwidth}
        \includegraphics[width=\textwidth]{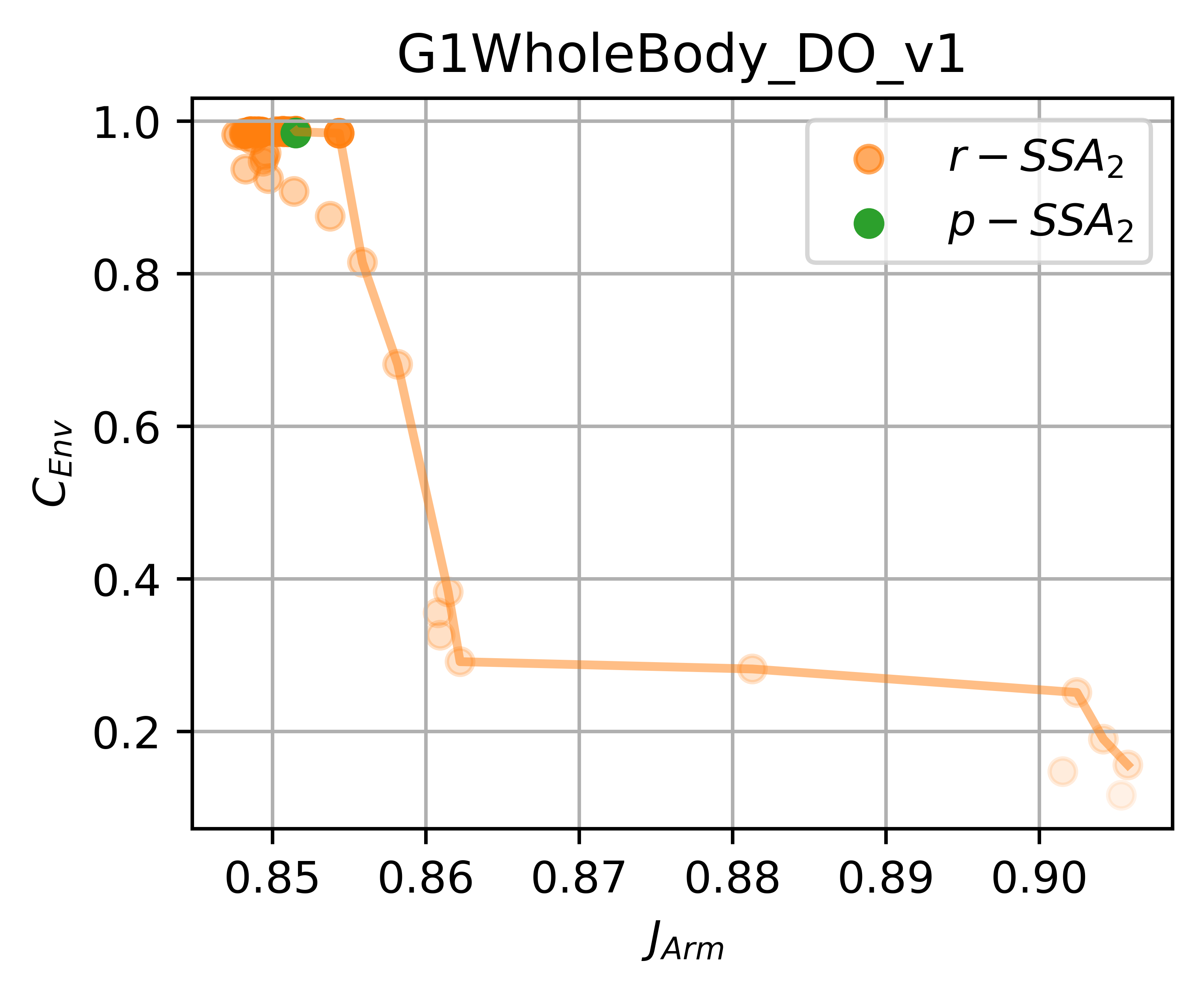}
        \caption{G1WholeBody Dynamic Obstalce V1}
        \label{fig:pareto_G1WholeBody_SG_DO_v1}
    \end{subfigure}
    \caption{Ablation study on r-SSA and p-SSA with G1WholeBody. Pareto fronts are plotted for r-SSA. r-SSA points are less transparent for larger $Q^{rssa}_s$.}
    \label{fig: ablation_study_whole}
\end{figure*}

\begin{figure*}[htbp]
    \centering
    
    \begin{subfigure}[b]{0.24\textwidth}
        \includegraphics[width=\textwidth]{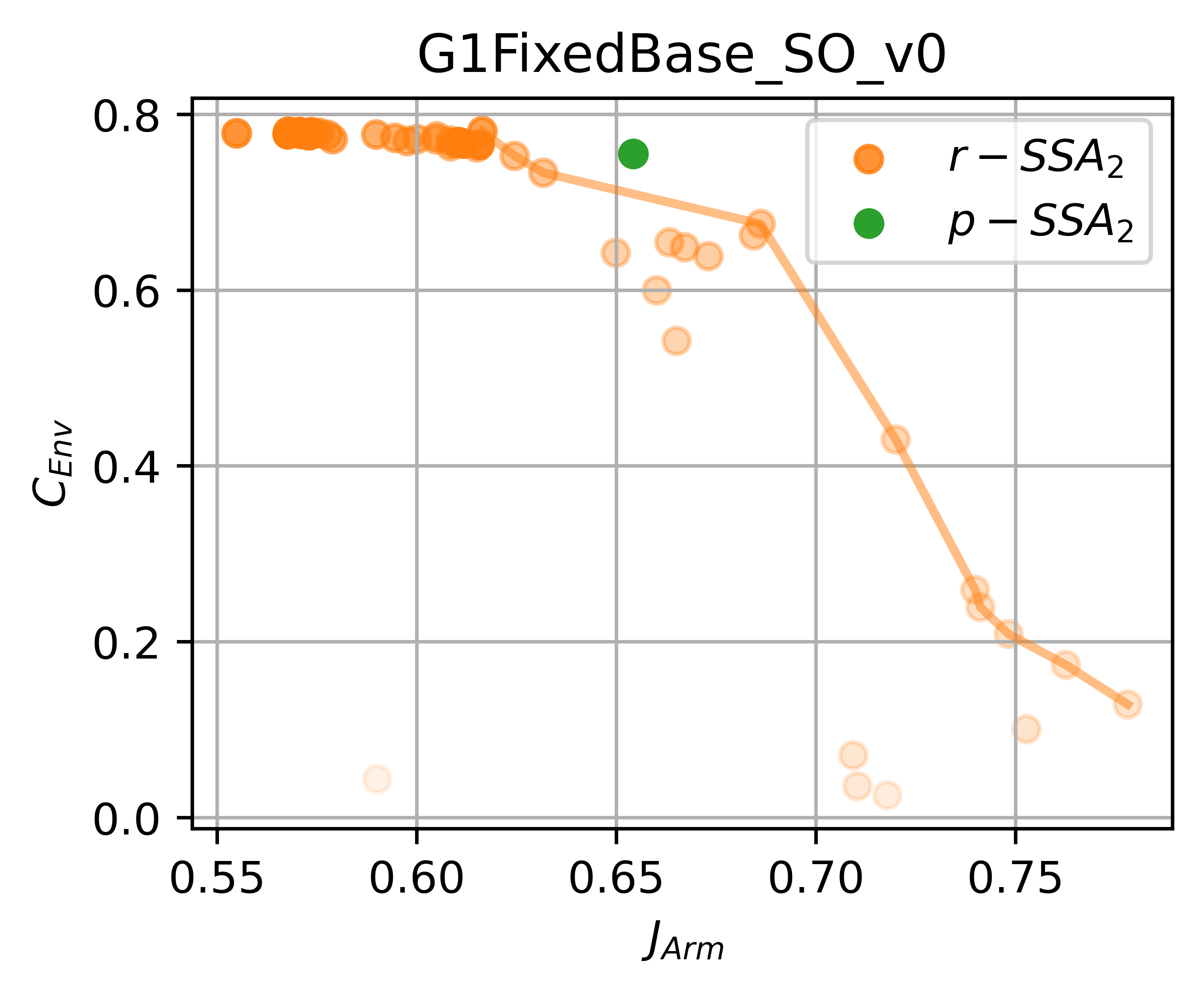}
        \caption{G1FixedBody Static Obstalce V0}
        \label{fig:pareto_G1FixedBase_SG_SO_v0}
    \end{subfigure}
    \begin{subfigure}[b]{0.24\textwidth}
        \includegraphics[width=\textwidth]{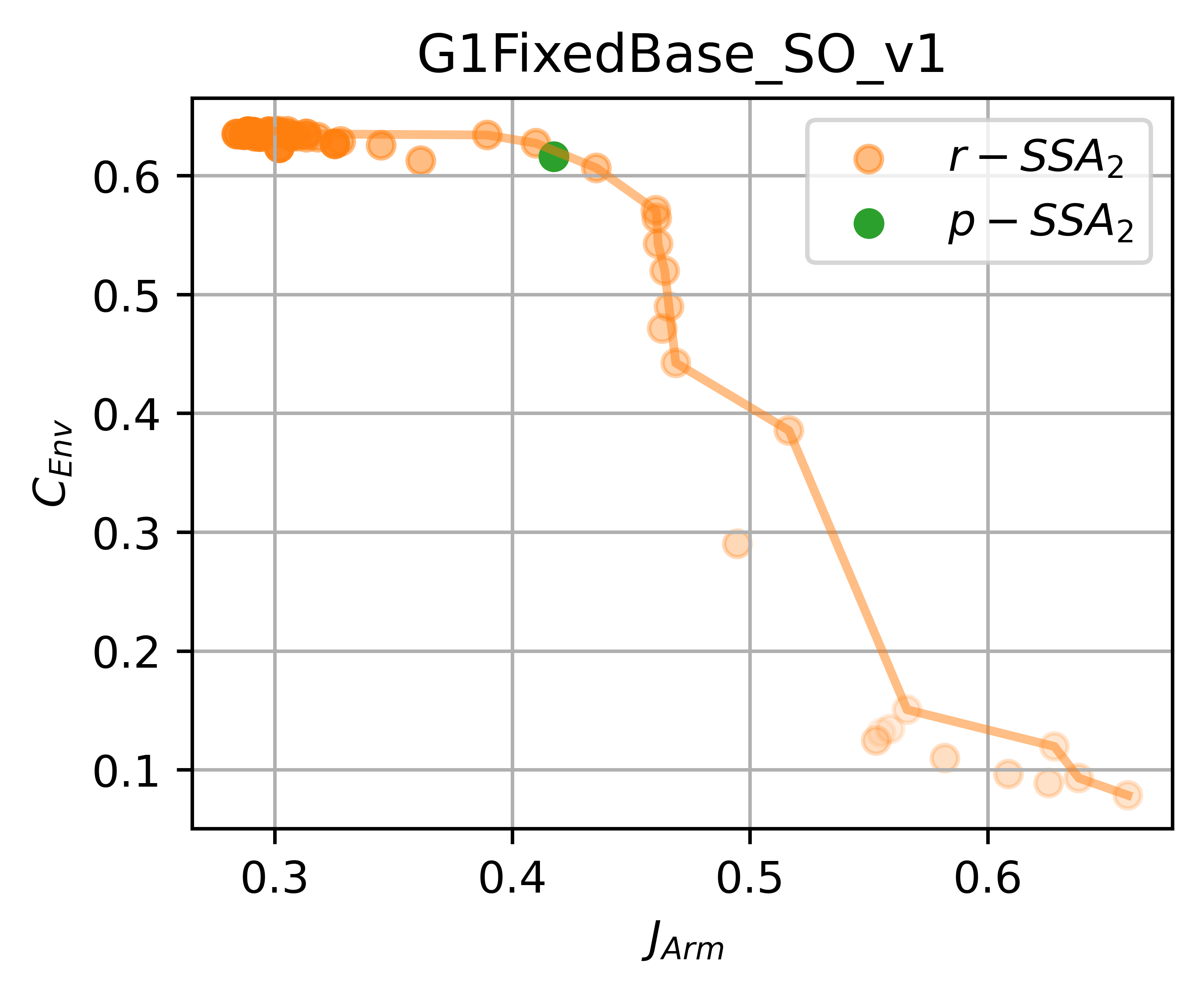}
        \caption{G1FixedBody Static Obstalce V1}
        \label{fig:pareto_G1FixedBase_SG_SO_v1}
    \end{subfigure}
    \begin{subfigure}[b]{0.24\textwidth}
        \includegraphics[width=\textwidth]{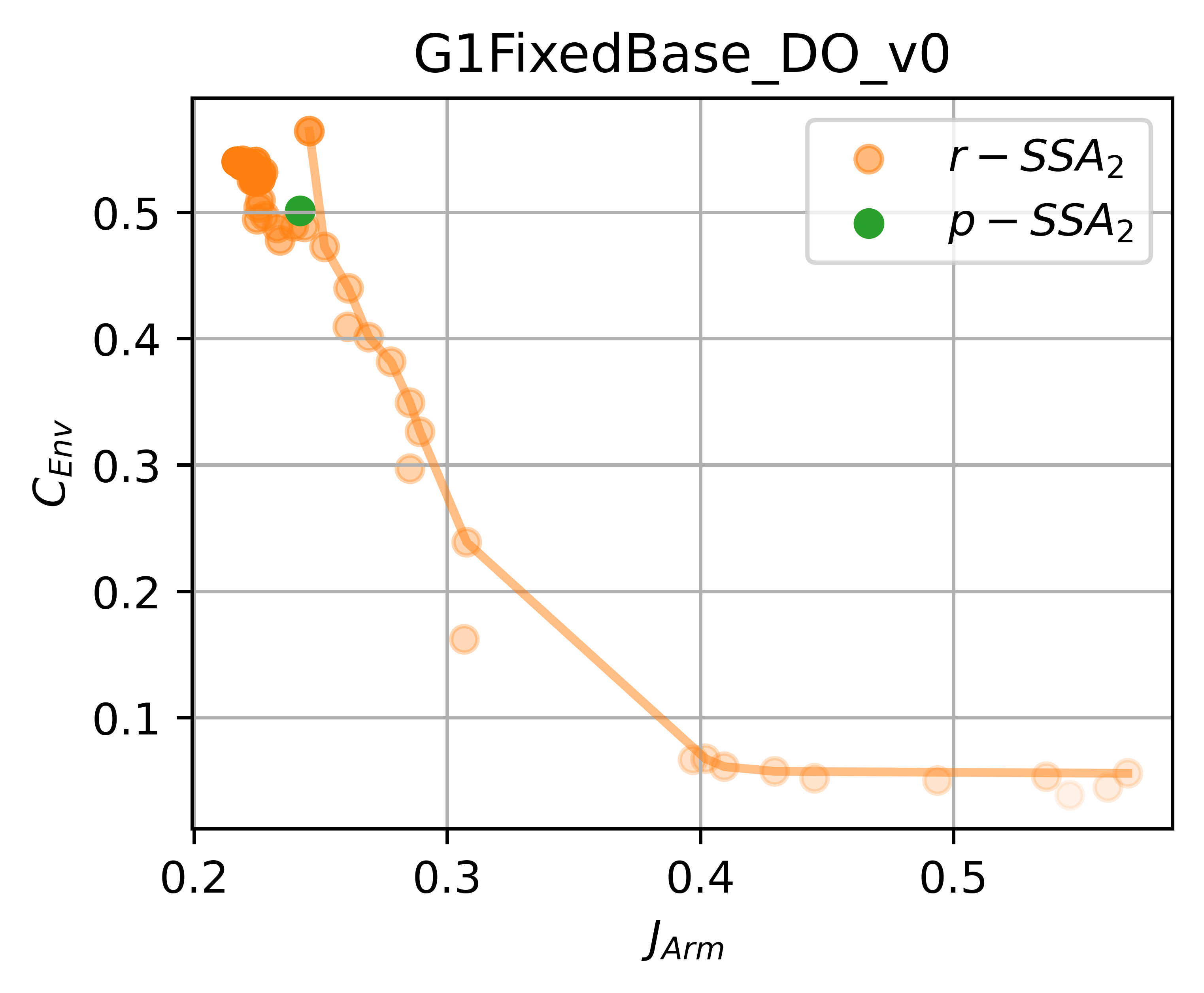}
        \caption{G1FixedBody Dynamic Obstalce V0}
        \label{fig:pareto_G1FixedBase_SG_DO_v0}
    \end{subfigure}
    \begin{subfigure}[b]{0.24\textwidth}
        \includegraphics[width=\textwidth]{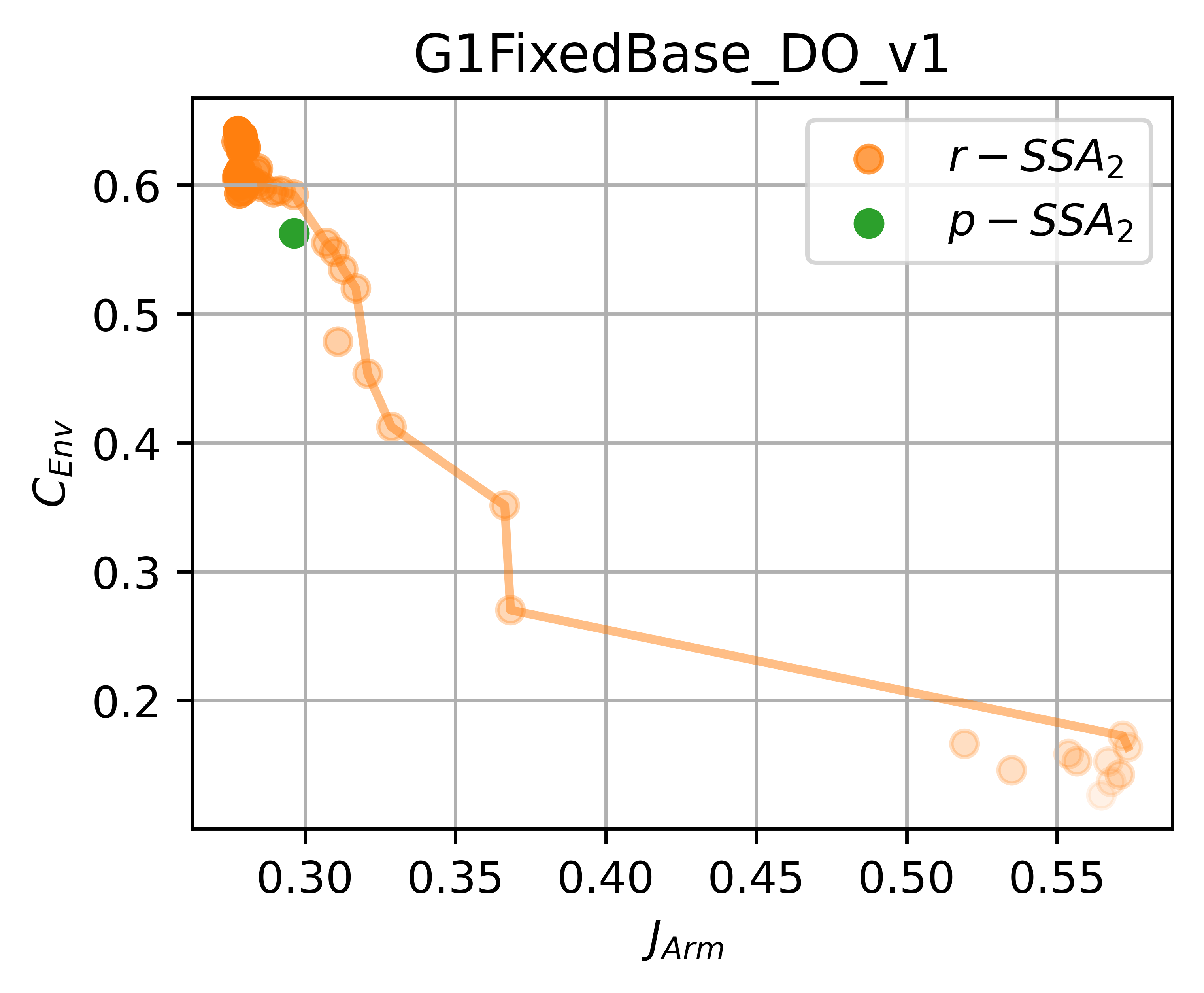}
        \caption{G1FixedBody Dynamic Obstalce V1}
        \label{fig:pareto_G1FixedBase_SG_DO_v1}
    \end{subfigure}
    \caption{Ablation study on r-SSA and p-SSA with G1FixedBody. Pareto fronts are plotted for r-SSA. r-SSA points are less transparent for larger $Q^{rssa}_s$.}
    \label{fig: ablation_study_fixed}
\end{figure*}

\subsection{Ablations on Relaxation Methods}

In this section, we perform ablation on r-SSA to see different performance-safety trade-offs and compare to p-SSA.
To achieve that, we benchmark r-SSA again on all eight tasks with a comprehensive range of $Q_s^{rssa}$ values: $\lambda I$ for $\lambda=a\cdot 10^b$, $a\in[1,9],b\in[0,5]$.
We select $C_\mathrm{env}$ as the proxy for safety and $J_\mathrm{Arm}$ as the proxy for performance.
See \Cref{fig: ablation_study_whole} and \Cref{fig: ablation_study_fixed} for the result.
Each orange point represents the $(J_\mathrm{Arm}, C_\mathrm{env})$ scores for r-SSA with a specific $Q_s^{rssa}$ value.
Pareto fronts are plotted for r-SSA.
A point with some $Q_s^{rssa}$ value is on the front if it is Pareto-optimal, meaning that there is no other ${Q_s^{rssa}}'$ value that can improve r-SSA in both metrics over $Q_s^{rssa}$.
Hence, the pareto front represents the optimal performance-safety trade-off curve for each task.

In all tasks, we find that when $Q_s^{rssa}$ is small, r-SSA would allow large control constraint violations to gain tracking performance (e.g., high $J_\mathrm{Arm}$ ranges).
As $Q_s^{rssa}$ becomes larger, the tracking performance starts to degrade since the control constraints are respected more, and the constraint satisfaction score improves.
At some point,the constraint satisfaction may stop improving while the task performance still degrades (e.g., lower $J_\mathrm{Arm}$ ranges in \cref{fig:pareto_G1WholeBody_SG_DO_v0}, \cref{fig:pareto_G1FixedBase_SG_SO_v0} and \cref{fig:pareto_G1FixedBase_SG_SO_v1}).
Based on the pareto front plots, one would prefer some $Q_s^{rssa}$ value in the top safety performance region that has the highest tracking performance.
Such tuning, however, would be tedious and cannot generalize to different tasks because the pareto front varies a lot for different tasks.
On the contrary, p-SSA automatically secures the sweet spot on the pareto fronts for all tasks.
Hence, answering \textbf{Q3}, while r-SSA can achieve optimal performance-safety trade-off with careful parameter tuning, p-SSA automatically achieves that with zero parameter tuning.


\begin{figure*}[ht]
\centering
    \includegraphics[width=1.0\linewidth]{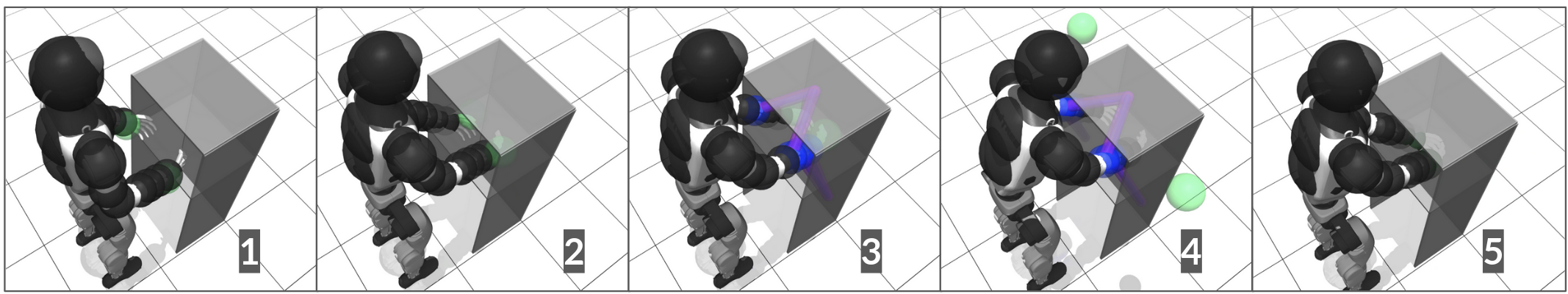}
    \caption{Safe teleopration with simluated Unitree G1 humanoid. The humanoid tracks wrist position goals (green) sent by the tele-operator while avoiding collision with the cabinet modeled by a few planes (gray). }
\label{fig: sim_g1_cabinet}
\end{figure*}

\subsection{Safe Teleoperation on Unitree G1 Humanoid}

To demonstrate the practical application of p-SSA, we perform a safe teleoperation task with a Unitree G1 humanoid.
During tele-operation, the humanoid motions may not exactly match the expectation of the operator due to various factors such as imperfect motion retargeting, delay, and human errors.
A safety filter like p-SSA would be useful in preventing unwanted collisions and ease the burden of safety from the operator.
In specific, we consider a scenario where the operator tele-operates the humanoid to organize a cabinet with tight opening.
The humanoid needs to put both arms inside the cabinet to manipulate objects without self-collision or colliding with the cabinet with any of its body parts.
The robot needs to avoid multiple obstacles (e.g., sides of the cabinet) at the same time in a confined space, yielding a typical dexterous safety problem in cluttered environments.

To generate $u_\mathrm{ref}$, we implemented a tele-operation framework that projects human wrist locations in human waist frame to those in humanoid frames.
A nominal controller solves inverse kinematics (IK) to acquire the desired upper body joint positions, and generates joint velocity commands via PID without considering safety.
We implement p-SSA with the G1FixedBase model described in \cref{sec:robot_model}.
Collision volumes of the cabinet are modeled as planes.
The design of energy functions $\phi_i$ respects the same safety margins in \cref{sec:exp_setting}.
Human motions, humanoid locations and obstacle locations are all sensed by an Apple Vision Pro (AVP) wore by the operator.

We show this task with both a simulated robot in MuJoCo and a real G1 humanoid with the same AVP-based teleoperation framework.
In both cases, the operator is encouraged to perform risky actions to see how p-SSA kicks in.
In simulation (see \cref{fig: sim_g1_cabinet}), we can see that p-SSA effectively rejects unsafe nominal controls.
For instance, even when the tele-operation goals travel through the cabinet walls, the robot arms stay inside the cabinet instead of following the goals blindly.
With real hardware (see \cref{fig:teaser}), we observe that p-SSA is able to prevent collisions with multiple types of improper operator actions.
In both simulation and real experiments, partial control constraints have to be relaxed, as indicated by the purple lines connecting collision pairs.
Hence, answering \textbf{Q4}, p-SSA is able to mitigate infeasible QPs in practical scenarios, making itself a reliable approach to the problem of dexterous safety in cluttered environments.





\section{Limitations}
\label{sec:limitation}

Both p-SSA and r-SSA demonstrated promising results when mitigating infeasible safe control optimizations for humanoids by relaxing the constraints with a minimal amount.
However, as long as relaxation is needed, the safety violation cannot be bounded, preventing any safety guarantee to hold.
To enhance our work with certain guarantees, it may be possible to adapt the original control constraints based on the solved minimal relaxation.
Then, we might be able to temporarily restore safety guarantees, enhancing post-infeasible-QP safety.

\section{Discussions and Conclusions}
\label{sec:conclusion}

In this paper, we proposes the Relaxed Safe Set Algorithm (r-SSA) and Projected Safe Set Algorithm (p-SSA) to mitigate infeasible safe control optimizations in the problem of dexterous safety in cluttered environments.
We showed that there are multiple sources of infeasibility, making it extremely hard to synthesize permanently feasible safe controllers for multi-constraint cases.
Through both simulation and real experiments on a Unitree G1 humanoid, we validated both r-SSA and p-SSA as practical approaches to computing safe control in challenging collision avoidance tasks.
We also showed that p-SSA achieved the optimal performance-safety trade-off across various tasks with zero parameter tuning.
For future work, we are interested in investigating optimal ways to balance between different safety constraints for p-SSA and improve the safety index design to account for higher order dynamics.

\section{Acknowledgement}

This work is supported by the National Science Foundation under grant No. 2144489.

We thank Zhixuan Liu for the help in making figures.


\bibliographystyle{plainnat}
\bibliography{references}

\newpage
\onecolumn
\section{Appendix}

\subsection{Self-collision Configuration}\label{append:self_collision}

For self-collision, we consider 29 pairs of humanoid bodies located at the joints listed below.

\begin{table}[h]
    \centering
    \begin{tabular}{|c|c|}
        \hline
        \textbf{Joint 1} & \textbf{Joint 2} \\
        \hline
        left\_shoulder\_roll\_joint & left\_elbow\_joint \\
        left\_shoulder\_roll\_joint & right\_shoulder\_roll\_joint \\
        left\_shoulder\_roll\_joint & right\_elbow\_joint \\
        left\_shoulder\_roll\_joint & L\_ee \\
        left\_shoulder\_roll\_joint & R\_ee \\
        left\_shoulder\_roll\_joint & torso\_link\_3 \\
        left\_elbow\_joint & right\_shoulder\_roll\_joint \\
        left\_elbow\_joint & right\_elbow\_joint \\
        left\_elbow\_joint & L\_ee \\
        left\_elbow\_joint & R\_ee \\
        left\_elbow\_joint & torso\_link\_1 \\
        left\_elbow\_joint & torso\_link\_2 \\
        left\_elbow\_joint & torso\_link\_3 \\
        right\_shoulder\_roll\_joint & right\_elbow\_joint \\
        right\_shoulder\_roll\_joint & L\_ee \\
        right\_shoulder\_roll\_joint & R\_ee \\
        right\_shoulder\_roll\_joint & torso\_link\_3 \\
        right\_elbow\_joint & L\_ee \\
        right\_elbow\_joint & R\_ee \\
        right\_elbow\_joint & torso\_link\_1 \\
        right\_elbow\_joint & torso\_link\_2 \\
        right\_elbow\_joint & torso\_link\_3 \\
        L\_ee & R\_ee \\
        L\_ee & torso\_link\_1 \\
        L\_ee & torso\_link\_2 \\
        L\_ee & torso\_link\_3 \\
        R\_ee & torso\_link\_1 \\
        R\_ee & torso\_link\_2 \\
        R\_ee & torso\_link\_3 \\
        \hline
    \end{tabular}
    \caption{29 Self-collision Humanoid Body Pairs}
    \label{tab:self_collision_pairs}
\end{table}

\subsection{Derivation of Safe Control Constraints}\label{append:safe_control_constraint}

We provide an example derivation of $L_f\bphi(x)$ and $L_g\bphi(x)$ for completing the control constraints in \eqref{eq:safe_contr_rssa} for r-SSA and \eqref{eq:safe_contr_pssa_phase_1} and \eqref{eq:safe_contr_pssa_phase_2} for p-SSA.
Assume that we have $M_\mathrm{env}$ different energy functions $[\phi^\mathrm{env}_1,\dots,\phi^\mathrm{env}_{M_\mathrm{env}}]$ for humanoid-obstacle collision and another $M_\mathrm{self}$ for self-collision: $[\phi^\mathrm{self}_1,\dots,\phi^\mathrm{self}_{M_\mathrm{self}}]$.
$\phi^\mathrm{env}_i=d_\mathrm{min,env}-d_i$ requires the distance between a collision body $j$ on the humanoid and an obstacle $k$ to be no less than $d_\mathrm{min,env}$.
Likewise, $\phi^\mathrm{self}_{i} = d_\mathrm{min,self}-d_i$ requires collision body $j$ and collision body $k$ to be at least $d_\mathrm{min,self}$ apart.
$d_i$ represents the current distance of interest with humanoid state $x$.
The final safety index is
\begin{equation}    
\bphi\defeq[\phi^\mathrm{env}_1,\dots,\phi^\mathrm{env}_{M_\mathrm{env}},\phi^\mathrm{self}_1,\dots,\phi^\mathrm{self}_{M_\mathrm{self}}]\in\RR^M
\end{equation}

We first derive $L_f\bphi(x)[i]$ and $L_g\bphi(x)[i]$ ($i^\mathrm{th}$ row) for $i\in[1,M_\mathrm{env}]$.
Assume that both collision bodies and obstacles are modeled as spheres with radius $R$, the distance $d_i$ is given by
\begin{equation}
    d_i = \|\bF_j(x) - t^O_k\| - 2R
\end{equation}
where $\bF_j(\cdot)$ is the forward kinematics function that computes the 3D center position of the collision sphere $j$ on the humanoid from joint position $x$.
$t^O_k$ is the 3D center position of the obstacle $k$.
We have
\begin{align}
    \frac{\partial d_i}{\partial x} &= \|\bF_j(x) - t^O_k\|^{-1}\left(\bF_j(x) - t^O_k\right)^\top \bJ_j(x)
\end{align}
where $\bJ_j(\cdot)$ is the Jacobian of the 3D center position of collision body $j$ with respect to the humanoid joint positions.
Then, we have
\begin{align}
    L_f\bphi(x)[i] &= \frac{\partial\bphi}{\partial x}f(x)[i] = \frac{\partial\phi^\mathrm{env}_i}{\partial x}f(x) = - \frac{\partial d_i}{\partial x}f(x) \\
    &= -\|\bF_j(x) - t^O_k\|^{-1}\left(\bF_j(x) - t^O_k\right)^\top \bJ_j(x) f(x) \label{eq:lf_env}
\end{align}
Similarly,
\begin{align}
    L_g\bphi(x)[i] &= -\|\bF_j(x) - t^O_k\|^{-1}\left(\bF_j(x) - t^O_k\right)^\top \bJ_j(x) g(x) \label{eq:lg_env}
\end{align}

To derive for $i\in[M_\mathrm{env}+1,M_\mathrm{env}+M_\mathrm{self}]$, we have
\begin{equation}
    d_i = \|\bF_j(x) - \bF_k(x)\| - 2R
\end{equation}
with gradient
\begin{align}
    \frac{\partial d_i}{\partial x} &= \|\bF_j(x) - \bF_k(x)\|^{-1}\left(\bF_j(x) - \bF_k(x)\right)^\top \left(\bJ_j(x) - \bJ_k(x)\right)
\end{align}
Similar to \eqref{eq:lf_env} and \eqref{eq:lg_env}, we have
\begin{align}
    L_f\bphi(x)[i] &= -\|\bF_j(x) - \bF_k(x)\|^{-1}\left(\bF_j(x) - \bF_k(x)\right)^\top \left(\bJ_j(x) - \bJ_k(x)\right) f(x) \\
    L_g\bphi(x)[i] &= -\|\bF_j(x) - \bF_k(x)\|^{-1}\left(\bF_j(x) - \bF_k(x)\right)^\top \left(\bJ_j(x) - \bJ_k(x)\right) g(x)
\end{align}


\subsection{Comparison of $\phi$ in Simulated Safe Goal Reaching}\label{append:phi_compare}

\begin{figure}[h]
    \centering
    \includegraphics[width=0.8\linewidth]{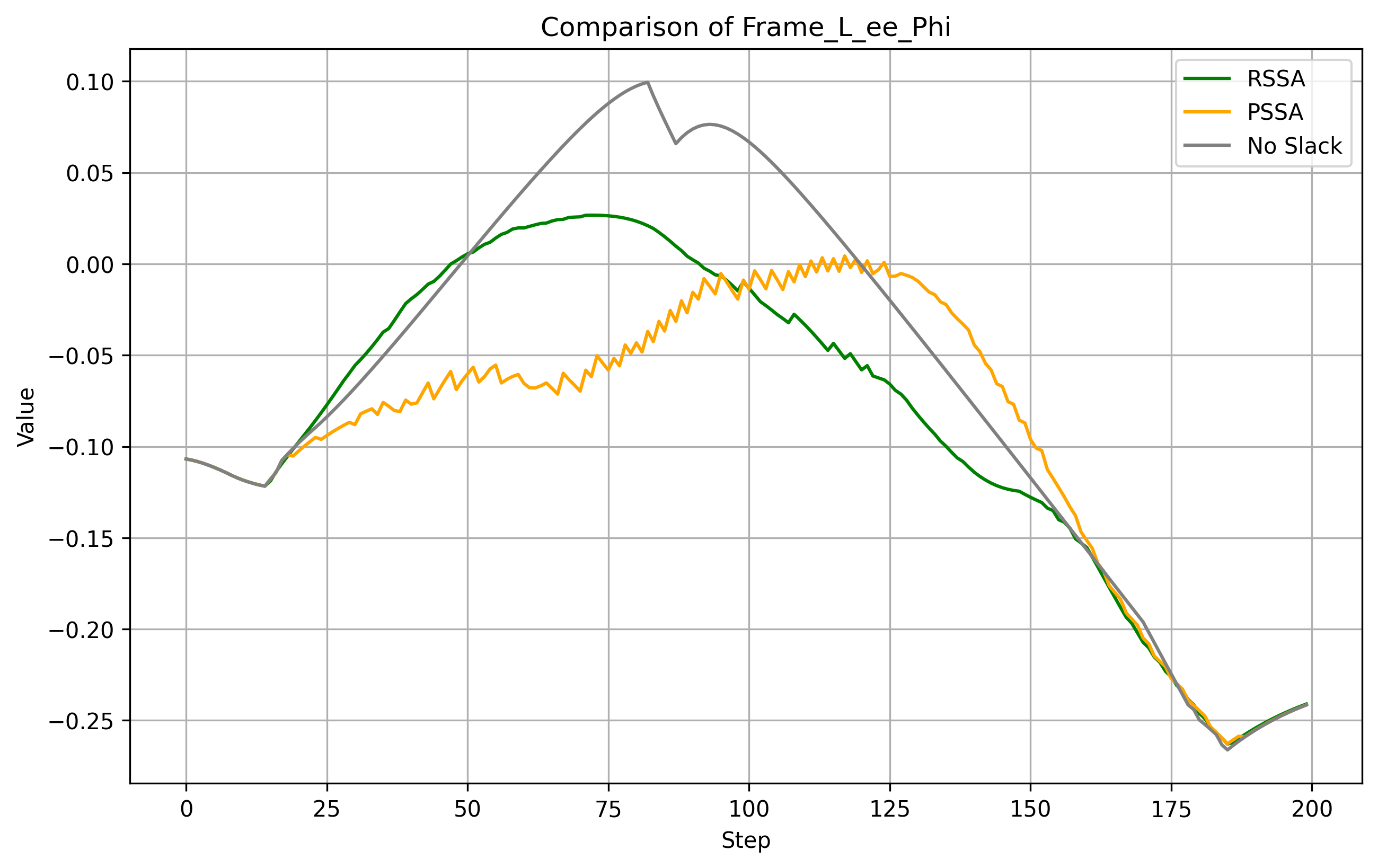}
    \caption{Change of $\phi$ that describes safety between the left hand and the nearest obstacle in \cref{fig: sim_example}. We can see that p-SSA consistently bounds $\phi$ below zero with some tolerance due to discretization. r-SSA allows more violations when handling infeasible QP. Naive SSA (No Slack) leads to the most significant violations.}
    \label{fig:l_ee_phi}
\end{figure}

\begin{figure}[h]
    \centering
    \includegraphics[width=0.8\linewidth]{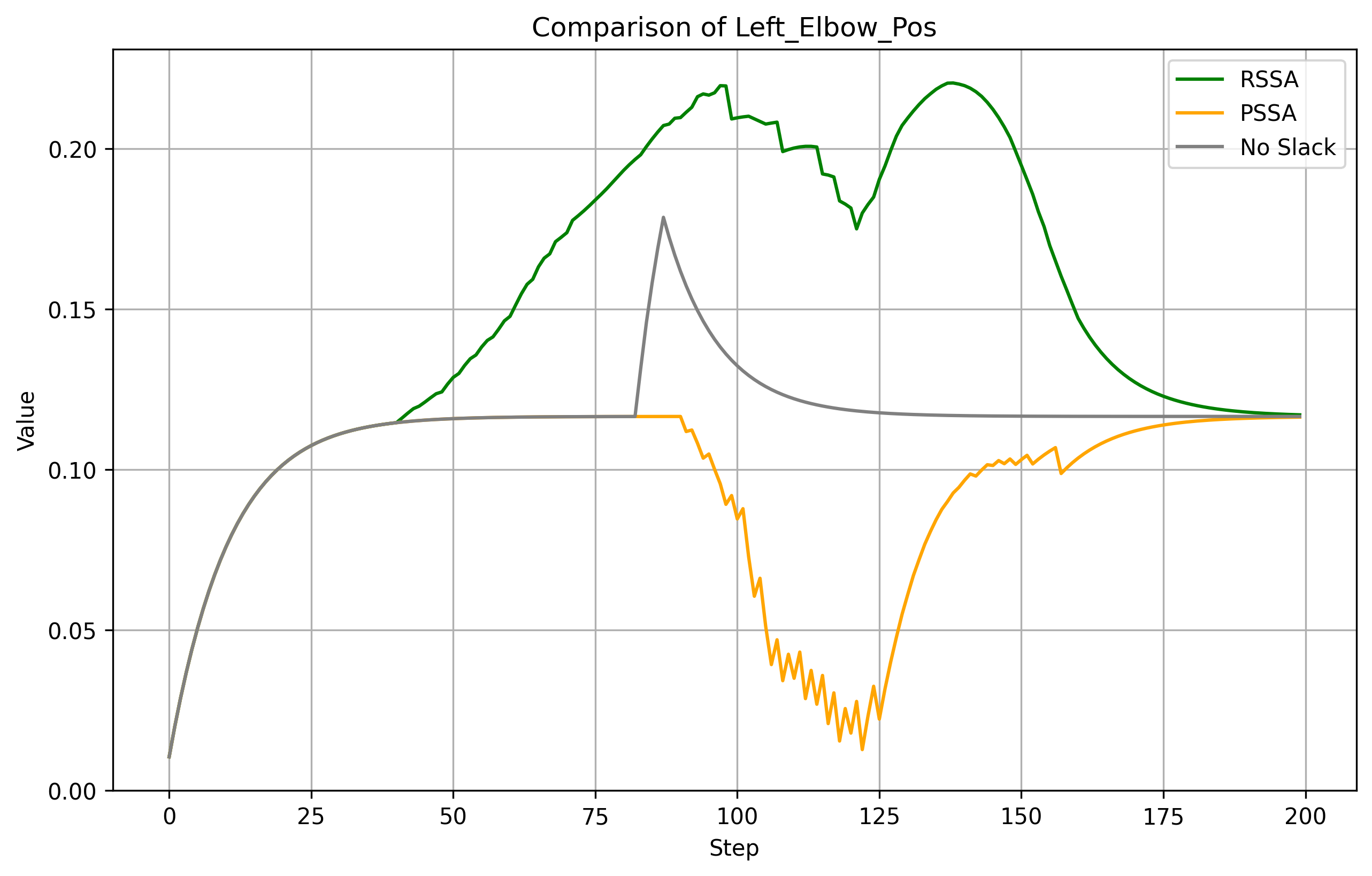}
    \caption{Change of left elbow joint position in \cref{fig: sim_example}.}
    \label{fig:l_elbow_pos}
\end{figure}

\begin{figure}[h]
    \centering
    \includegraphics[width=0.8\linewidth]{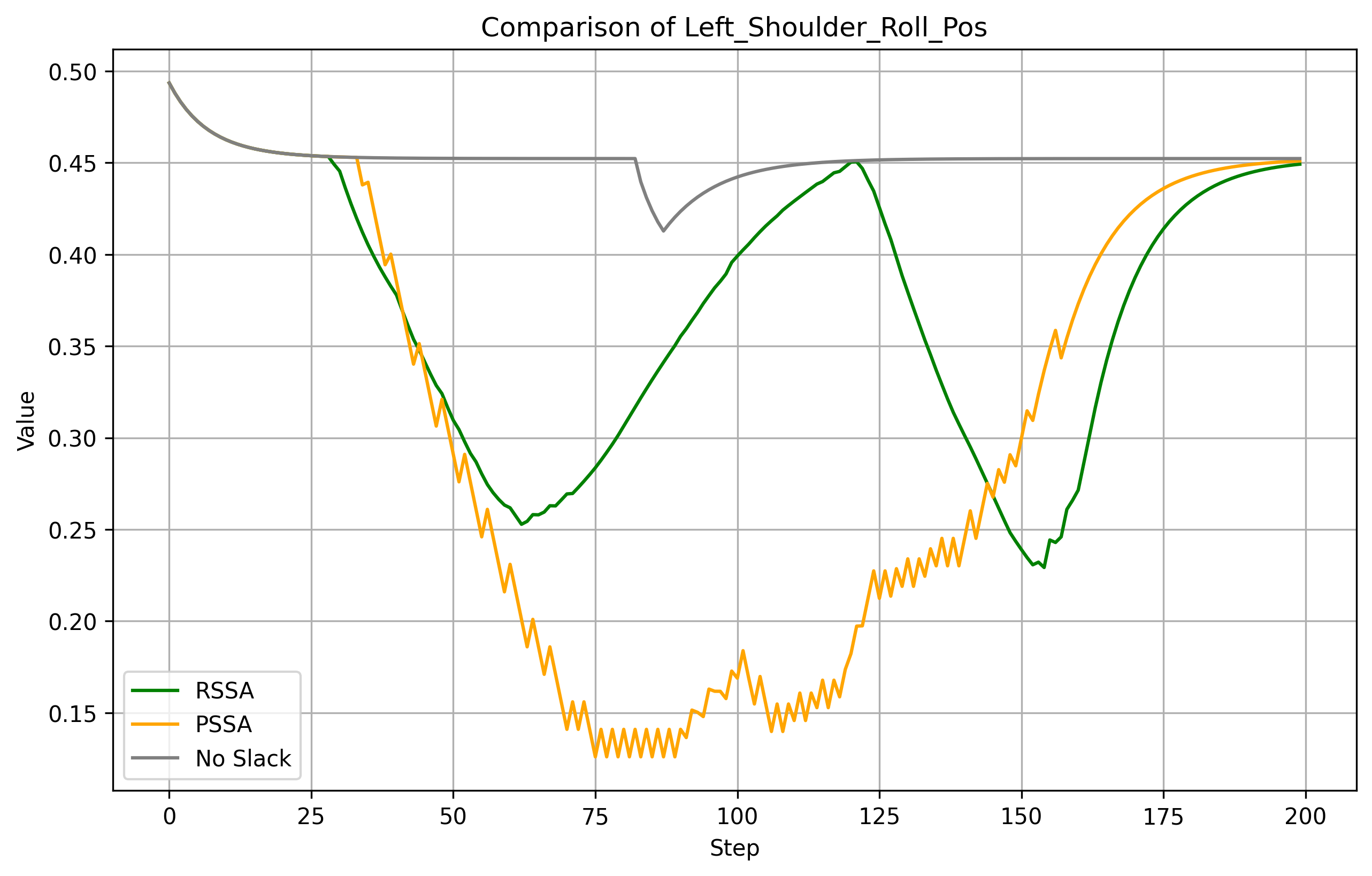}
    \caption{Change of left shoulder roll joint position in \cref{fig: sim_example}.}
    \label{fig:l_shoulder_roll}
\end{figure}

\end{document}